\documentclass[accepted]{uai2026} % after acceptance, for a revised version; 
\usepackage[american]{babel}
\usepackage{natbib} % has a nice set of citation styles and commands
\usepackage{mathtools} % amsmath with fixes and additions
\usepackage{booktabs} % commands to create good-looking tables
\usepackage{tikz} % nice language for creating drawings and diagrams
\usepackage{amsmath}
\usepackage{amssymb}
\usepackage{amsthm}
\usepackage{multirow}
\usepackage[most]{tcolorbox}
\usepackage{algorithm}
\usepackage{algpseudocode}

\newcommand{\diag}[1]{\operatorname{diag}(#1)}
\newcommand{\A}{\mathbf{A}}
\newcommand{\B}{\mathbf{B}}
\newcommand{\W}{\mathbf{W}}
\newcommand{\s}{\mathbf{s}}
\newcommand{\h}{\mathbf{h}}
\newcommand{\U}{\mathbf{U}}
\newcommand{\V}{\mathbf{V}}
\newcommand{\D}{\mathcal{D}}
\newcommand{\N}{\mathcal{N}}
\newcommand{\y}{\mathbf{y}}
\newcommand{\x}{\mathbf{x}}
\newcommand{\eps}{\boldsymbol{\epsilon}}
\newcommand{\btheta}{\boldsymbol{\theta}}
\newcommand{\R}{\mathbb{R}}

\title{Bayesian Adaptation Gym: A Benchmark for the Bayesian Low-Rank Adaptation of Multi-Modal Language Models}

\author[1]{\href{mailto:<colinsamplawski@gmail.com>?Subject=Your UAI 2026 paper}{Colin Samplawski}{}}
\author[1]{Ramneet Kaur}
\author[1]{Manoj Acharya}
\author[1]{Anirban Roy}
\author[1]{Adam D. Cobb}
\affil[1]{%
Neuro-Symbolic Computing and Intelligence Research Group\\
Computer Science Laboratory\\
SRI International
}
  
\begin{document}
\maketitle

\begin{abstract}
Large multi-modal language models are increasingly deployed in high-stakes domains, making well-calibrated uncertainty essential. Traditional Bayesian methods approximate posteriors over all model weights, which becomes intractable for modern large models. For this reason, recent work instead considers Bayesian low-rank adaptation to enable tractable posterior approximation. Due to a lack of a standardized benchmark to evaluate these approaches, it remains unclear where these methods provide meaningful benefits. To fill this gap, we introduce Bayesian Adaptation Gym (BAG), a benchmark for the Bayesian adaptation of multi-modal language models. BAG provides reference implementations of classic Bayesian baselines and state-of-the-art adaptation methods, along with a multi-modal dataset and task suite designed to probe calibration, robustness under distribution shift, and decision-making under uncertainty via active learning. Using BAG, we conduct and report extensive experiments across model sizes, datasets, and tasks to highlight the successes and failures of current Bayesian adaptation approaches. To enable further research, BAG is fully open source: \url{https://github.com/SRI-CSL/BayesAdapt}.

\end{abstract}

\section{Introduction}
The uncertainty quantification (UQ) of large, multi-modal language models has become an increasing priority of the research community due to widespread use of these models in many high-stakes domains, such as healthcare \citep{poon2025adoption}, scientific research \citep{chen2025ai4research}, or even everyday use by the general public. UQ offers a promising way to ensure well-calibrated and trustworthy outputs from these models which are known to hallucinate and output unreliable information \citep{kalai2025language}.

A broad family of approaches has emerged in the space of UQ over large models. The most popular solutions are ``black box`` approaches, such as self consistency-based approaches \citep{wang2022self} or verbalized confidence \citep{xuan2025seeing}, which consider only the uncertainty in the output space of the model. While often effective, they are incapable of probing the uncertainty in the parameters of the model itself. For this, we require Bayesian methods \citep{wang2020survey}. Here, uncertainty quantification is performed by directly approximating a distribution over the weights of the model $P(\W|\D)$ where $\mathcal{D}$ is a training (or fine-tuning) dataset, and $\W$ are the model parameters. We can then estimate a model's output uncertainty by marginalizing over this parameter posterior distribution via a Bayesian model average.

The fundamental challenge is then finding a high quality approximation of the parameter posterior $P(\W|\D)$. It is well known that this quickly becomes intractable as the dimensionality of $\W$ grows, making the huge parameter sizes of modern LLMs especially challenging. However, an emerging body of work has focused instead on the task of Bayesian adaptation \citep{lap,blob,tfb,scalabl,pham2025promoting}. In this setting, we model the posterior distribution not over the full parameters $\W$, but over low-rank adaptation (LoRA) parameters learned during fine-tuning. Due to the low-rank nature of these parameters, approximating this LoRA posterior remains tractable.

The current benchmarking protocol for the Bayesian adaptation of LLMs used throughout prior work originates from the Laplace LoRA work of \cite{lap}. Under this setup, models are fine-tuned and tested on multiple-choice commonsense reasoning and trivia datasets. Classification accuracy, along with uncertainty metrics such as expected calibration error (ECE) and negative log likelihood (NLL) are used to judge the performance of different techniques. This protocol has become the standard evaluation scheme used in essentially all subsequent work on this problem \citep{blob,tfb,scalabl,pham2025promoting}.

A major limitation of this protocol is that for most of the datasets considered, there is little ``headroom'' for fine-tuning. By this we mean that there is a small performance difference before and after fine-tuning (see Table \ref{tab:headroom}). 
This can happen either because the training set provides limited transferable signal, and/or because the base model already has strong performance due to pretraining exposure (implicit or explicit). 
This low headroom often makes the Bayesian adaptation process less of a fine-tuning and more of a very expensive and complex in-distribution calibration process. In Section \ref{sec:exp_calibration} we show that a simple temperature scaling baseline \citep{tempscale} is often competitive in this evaluation setup. Additionally, prior tasks are limited to text-based datasets and unimodal LLMs, ignoring current multi-modal use cases. 

Beyond dataset issues, prior work provides limited discussion of resource usage, despite Bayesian adaptation methods often incurring substantial additional cost. Moreover, evaluations are typically restricted to a narrow band of backbone sizes, most commonly 7-8B parameter LLMs, leaving it unclear how these methods scale in memory and latency with other model sizes. These problems are further intensified by the lack of a modular framework with which to build and evaluate methods.

To fill these gaps, we introduce Bayesian Adaptation Gym (BAG), a benchmark for the Bayesian adaptation of multi-modal language models. We design BAG along the following axes: first, we prioritize multi-modal tasks with meaningful adaptation headroom, where fine-tuning offers clear gains over the pretrained model. Second, we measure uncertainty quality and robustness for both in-distribution data and across a broader set of out-of-distribution shifts than in prior work. As such, we are able to highlight clearer differences in uncertainty performance across methods. Third, we include the decision-centric evaluation of active learning, where uncertainty directly affects which data are acquired and thus impacts label efficiency. Fourth, we treat resource utilization as a first-class metric by standardizing reporting of inference-time resource usage across a range of backbone sizes. 

Finally, BAG is released as a modular and extensible framework with reference implementations of classic Bayesian deep learning and recent Bayesian adaptation methods, enabling controlled and reproducible comparisons under a unified training and evaluation pipeline, while also making it straightforward to add novel datasets and adaptation techniques.

The contributions of our work are as follows:
\begin{itemize}
    \item We introduce Bayesian Adaptation Gym (BAG) a modular and extensible Python framework to benchmark Bayesian adaptation of VLMs (Section \ref{sec:bag}).
    \item We include reference implementations of classic and state-of-the-art techniques for the problem of Bayesian adaptation.
    \item BAG includes a dataset suite made up new multi-modal tasks, as well as new protocols for out-of-distribution evaluation and active learning.
    \item Using BAG, we perform and present an extensive set of experiments, where we explore performance across different model sizes, datasets, and tasks.
    \item To enable further research, BAG is fully open source: \url{https://github.com/SRI-CSL/BayesAdapt}
\end{itemize}

\begin{figure*}
    \centering
    \includegraphics[width=\linewidth]{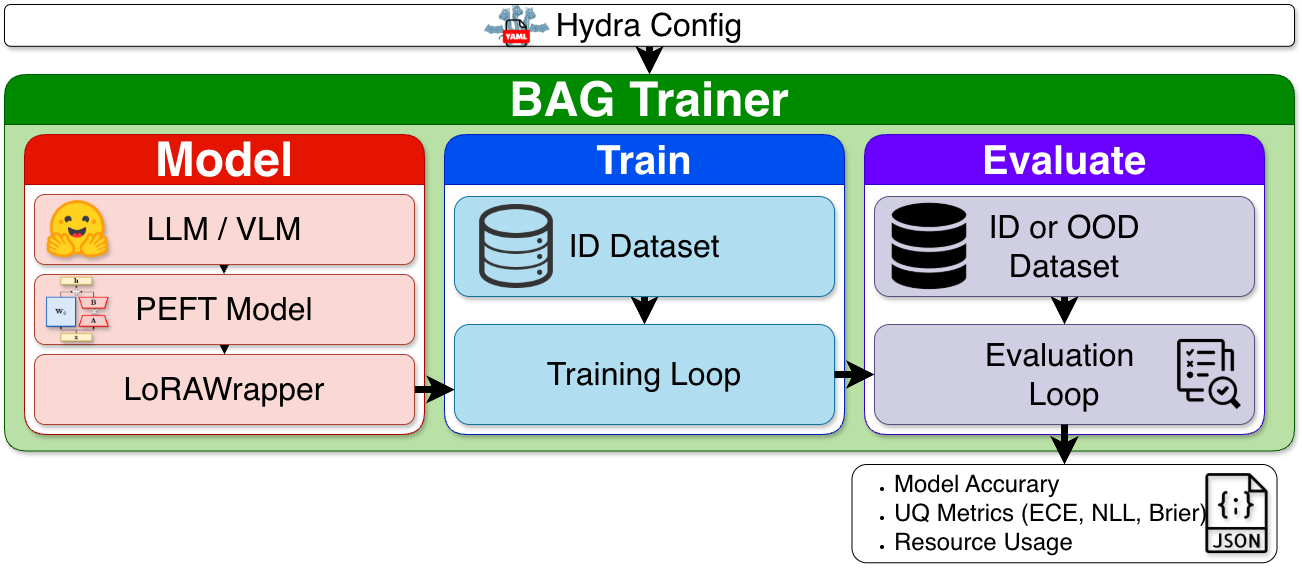}
    \caption{Overview of the BAG \texttt{Trainer} object. Every box represents a component in the framework which can be modified or extended.}
    \label{fig:bag_trainer}
\end{figure*}

\section{Prior Work and Preliminaries}
Prior work on uncertainty quantification of recent models has attracted a large literature of techniques to apply to this problem. There remains limited work which aims to rigorously compare these approaches. However, what does exist \citep{vashurin2025benchmarking} does not consider Bayesian methods. On this front, there have been a number of prior works which rigorously benchmark Bayesian techniques \citep{band2021benchmarking,wilson2022evaluating,ursabench}. However, these prior benchmarks originate from the pre-LLM era and are primarily concerned with the posterior approximation of image classification CNNs trained from scratch, leaving the increasingly important setting of Bayesian adaptation of pretrained VLMs without a standardized evaluation protocol.

\subsection{Low-Rank Adaptation}
First introduced by \cite{lora}, low-rank adaptation (LoRA) has become a common technique for the tractable fine-tuning of large models. Let $\W_0 \in \R^{n \times d}$ be the pretrained weights a single linear layer, where $d$ is the embedding dimension and $n$ is the output size. When fine-tuning with LoRA, the pretrained weights $\W_0$ are kept fixed and instead we learn a new pair of low-rank matrices $\A \in \R^{r \times d}$ and $\B \in \R^{n \times r}$, for LoRA rank $r \ll \min(n,d)$. We can then compute the forward pass as the addition of linear operations on a batch of $B$ input features $\mathbf{x} \in \mathbb{R}^{B \times d}$:
\begin{align}
    \h = \x\W_0^T + \x (\B \A)^T
\end{align}
The low-rank nature of $\A$ and $\B$ lead to considerable memory savings during fine-tuning compared to full weight updates.

\subsection{Bayesian Low-Rank Adaptation}
Bayesian low-rank adaptation replaces the point-estimate adapter of a typical LoRA fine-tuning with a distribution over adapter parameters. This treats $\A$ and/or $\B$ as random variables with a prior and posterior inferred using a fine-tuning dataset $\D$. More concretely, for frozen pretrained base weights $\W_0$ and LoRA parameters $\Delta \coloneqq (\A,\B)$, the Bayesian predictive distribution marginalizes over the adapter posterior:
\begin{equation}
P(\y \mid \x, \D) = \int P(\y \mid \x, \W_0, \Delta)\, P(\Delta \mid \D)\, \mathrm{d}\Delta,
\end{equation}
which is approximated in practice using a tractable posterior approximation.
Since $\Delta$ lives in a low-dimensional subspace determined by the LoRA rank $r$, this posterior approximation and inference is far more tractable than full Bayesian inference over $\W_0$, while still capturing parameter uncertainty in the full weight space.
In practice, we approximate this intractable integral by taking the Bayesian model average over $N$ Monte Carlo samples from the posterior approximation.

\begin{table*} 
\centering
\caption{Example of dataset headroom on prior datasets and datasets introduced with BAG (underlined).}
\begin{tabular}{@{}cccccccc@{}}
\toprule
\textbf{Metric} & \textbf{Method} & \textbf{ARC-Challenge} & \textbf{Winogrande (s)} & \underline{\textbf{Circuit Logic}} & \underline{\textbf{SLAKE}} & \underline{\textbf{SRQA}} & \underline{\textbf{MathVerse}} \\
\midrule
\multirow{2}{*}{\textbf{ACC ($\uparrow$)}}
& Pretrained & 0.912 & 0.511 & 0.387 & 0.789 & 0.673 & 0.502 \\
& MLE      & 0.918 & 0.760& 0.842 & 0.892 & 0.966 & 0.513 \\
\midrule
\multirow{2}{*}{\textbf{NLL ($\downarrow$)}}
& Pretrained & 0.947 & 1.656 & 2.007 & 0.778 & 1.918 & 1.749 \\
& MLE      & 0.833 & 2.614 & 0.294 & 0.980 & 0.092 & 4.757 \\
\bottomrule
\end{tabular}
\label{tab:headroom}
\end{table*}

\section{Bayesian Adaptation Gym} \label{sec:bag}
Bayesian Adaptation Gym (BAG) is a benchmark and programming framework for evaluating Bayesian low-rank adaptation of multi-modal language models. A core design goal is to make comparisons between methods controlled and reproducible by standardizing (i) how Bayesian adapters are inserted into the model, (ii) how evaluation is performed, (iii) how datasets and tasks are formatted, and (iv) how evaluation metrics (including resource costs) are measured and reported. BAG is also designed with an eye toward future research in this emerging area. Its modular and extensible nature makes it straightforward to test new VLM variants from HuggingFace (usually just one line change in the config file), new datasets, and new Bayesian adaptation techniques.

Loosely inspired by PyTorch Lightning \citep{Falcon_PyTorch_Lightning_2019}, the high level structure of BAG is built around a monolithic \texttt{Trainer} object with modular training and evaluation loops (Figure \ref{fig:bag_trainer}). The \texttt{Trainer} and its behavior are completely defined by a \texttt{hydra} configuration file \citep{Yadan2019Hydra}. The \texttt{Trainer} includes the boilerplate code needed to load any HuggingFace VLM and insert a LoRA adapter around every desired layer. A key abstraction of BAG is then a set of implemented \texttt{lorawrappers} which provide a modular way to define the logic of each Bayesian adaptation technique. BAG then defines a universal dataset output format for collation, allowing the easy integration of HuggingFace or custom datasets into the training and evaluation pipelines. The \texttt{Trainer} and \texttt{lorawrappers} implementations are both highly modular and extensible, allowing the straightforward modification of nearly every component either by modifying \texttt{hydra} configuration or extending to new subclasses. Through the use of the \texttt{ray} plugin \citep{ray} to \texttt{hydra}, BAG easily supports performing  massive parallel sweeps on multi-GPU nodes.

\subsection{Implemented Methods}
BAG offers implementations of standard Bayesian methods as well as recent state-of-the-art approaches designed specifically for the problem of Bayesian low-rank adaptation. We briefly describe them here, with more details for each method provided in Appendix Section \ref{sec:method_details}.

\textbf{MLE and MAP:} A fundamental baseline is the standard LoRA fine-tuning process itself, where the low-rank matrices are learned by gradient-based optimization without explicitly modeling a weight posterior. We refer to this as the maximum likelihood estimate (MLE). Following prior work, we also consider a Maximum A Posteriori (MAP) variant by adding weight decay during LoRA fine-tuning.

We then group the Bayesian methods into two categories: post-hoc methods, which require access to a pretrained adapter and approximate the posterior after training, and online methods, which minimize training loss and learn a posterior approximation end-to-end during fine-tuning.

\textbf{Laplace:} The Laplace-LoRA approach of \cite{lap} constructs a local Gaussian approximation of the posterior around a trained adapter using second-order curvature, with additional structure and approximations to keep the Hessian computations tractable.

\textbf{TFB:} Training-Free Bayesianization (TFB) \cite{tfb} converts a trained MLE adapter into a Bayesian one by restricting the posterior to a low-rank isotropic Gaussian over the adapter weights. Rather than optimizing variational parameters, TFB selects a single global noise scale $\sigma_q$ via a simple binary search which seeks to maximize uncertainty while keeping classification performance on an in-distribution ``anchor'' dataset within a tolerance $\epsilon$.

\textbf{TempScale:} Although not a Bayesian method, we also include temperature scaling \citep{tempscale} as a simple calibration baseline. Using gradient optimization on a validation set, we learn a single scalar $T$ which rescales the probabilities of a pretrained MLE adapter.

\textbf{Deep Ensembles:} Deep ensembles \citep{deepensembles} train $N$ independent LoRA adapters using different random seeds. In principle, this method samples weights from the true posterior, but comes at the significant cost of multiplying training and storage cost by $N$.

%\subsubsection{Variational Inference Methods}
%Variational inference methods posit a tractable family $q_{\theta}(W)$ and fit it by maximizing the evidence lower bound (ELBO).

\textbf{MCDropout:} The Monte Carlo (MC) dropout approach of \cite{mcdropout} reinterprets standard dropout as a form of approximate variational inference, leading to an implicit posterior approximation via Bernoulli masking. At test time, dropout remains enabled and $N$ stochastic forward passes are used to approximate the predictive distribution.

\textbf{BLoB:} The Bayesian LoRA by Backprop (BLoB) approach of \cite{blob} extends Bayes by Backprop \citep{bbb} to LoRA by asserting a variational diagonal Gaussian posterior over the LoRA $\A$ matrix, learning both a mean and variance parameter for each weight in $\A$. Using the reparameterization trick, we can draw stochastic weight samples during training to optimize the evidence lower bound (ELBO) \citep{elbo}, jointly estimating the adapter weights and their uncertainty during fine-tuning. %Similarly, at test we draw weight samples to compute a Bayesian model average.

\textbf{ScalaBL:} Similar to BLoB, the scalable Bayesian low-rank adaptation (ScalaBL) approach of \cite{scalabl} performs stochastic variational inference in an $r$-dimensional subspace (where $r$ is the LoRA rank), then repurposes the $\A$ and $\B$ matrices to map low-dimensional samples into the full-weight space. This reduces the number of additional variational parameters compared to BLoB while still remaining broadly performant.

\begin{figure*}[h!]
    \centering
    \includegraphics[width=\linewidth]{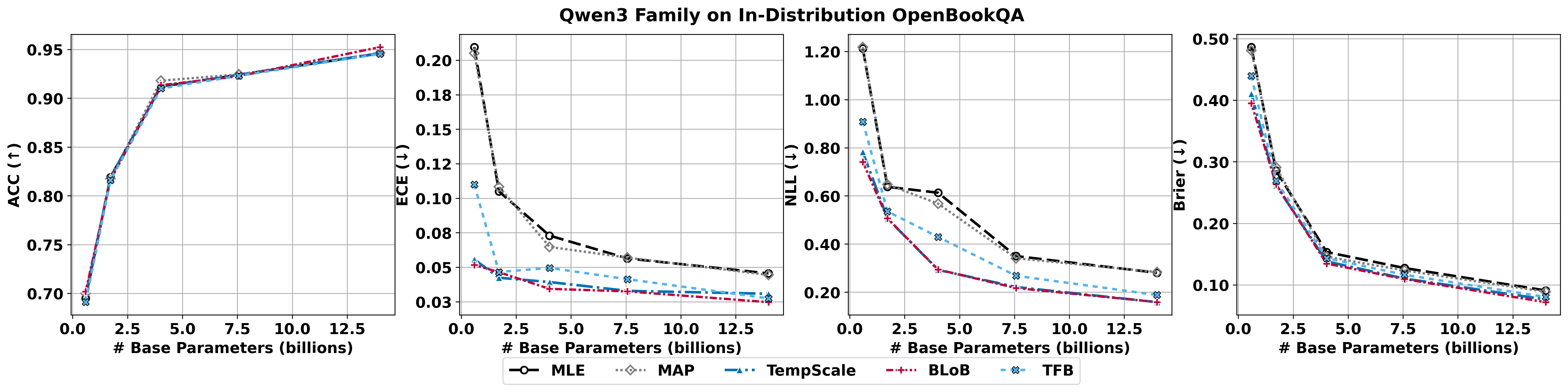}
    \caption{Test of model calibration using OpenBookQA dataset from prior work across a range of model sizes in the Qwen3 family. We see that simple temperature scaling is competitive baseline on this evaluation setup (which is often used in prior work).}
    \label{fig:obqa_id_calib}
\end{figure*}
\subsection{Datasets}\label{sec:data}
In BAG, we carry forward all text-only datasets typically used in prior work, namely: Winogrande \citep{winogrande}, ARC-Easy/Challenge \citep{arc}, OpenBookQA \citep{obqa}, BoolQ \citep{boolq}, and MMLU \citep{mmlu}. 

We also incorporate a  new text-only dataset derived from the Reasoning Gym benchmark of \cite{reasoninggym}. Specifically, we include the \textbf{Circuit Logic} task, where the model must determine the output truth value of a randomly generated Boolean circuit displayed in Unicode, given input assignments for each variable. Because each circuit can be represented equivalently as a logical formula, the task naturally supports controlled distribution shifts (e.g., training on one representation and evaluating on the other). This provides a useful stress test for out-of-distribution generalization and for whether uncertainty estimates increase appropriately when the representation changes while the underlying reasoning problem remains the same. Further details are provided in Appendix Section \ref{sec:circuit_logic}.

In an effort to move beyond a unimodal evaluation setup, we additionally add the following vision-based datasets:
\textbf{SLAKE} \citep{slake} is a medical visual question answering dataset built around radiology images, with questions that require both understanding the image (e.g., anatomy/abnormalities) and applying clinical knowledge. Further details are provided in Appendix Section \ref{sec:slake}.

\textbf{MathVerse} \citep{mathverse} is a mathematical reasoning benchmark where each problem pairs a natural language question with a visual input (e.g., geometric diagrams, plots, tables, etc.) to test whether a model can reason mathematically about what it sees. Further details are provided in Appendix Section \ref{sec:mathverse}.

\textbf{MMStar} \citep{mmstar} is a dataset of hand-selected image-question pairs which test a model's visual perception and understanding. It is designed specifically such that there is minimal leakage in pretrained models and that the image content must be understood to correctly answer the question. Further details are provided in Appendix Section \ref{sec:mmstar}.

A limitation of these datasets is their relatively small size. To address this, we introduce \textbf{SymbolicRegressionQA} (SRQA), a novel multi-modal dataset based on the task of symbolic regression. In symbolic regression, the goal is to recover a symbolic equation $y=f(x)$ for an unknown function $f$, given sampled numerical input–output pairs $(x,y)$.

We leverage the data generation code of \cite{snip}, which can be used to draw an effectively unlimited number of samples of symbolic expressions together with corresponding numerical input–output data. We configure the generator to produce only one-dimensional functions and render each sampled function as a plot using its associated $(x,y)$ samples. For every function, we additionally sample three distractor expressions, yielding a four-way multiple-choice question in which the model must identify the correct expression given the plot image. This construction has several appealing properties: it supports near-unlimited data samples, its difficulty is easily tuned (e.g., by adjusting the data generator parameters or increasing the number of answer choices), and it is unlikely to overlap with typical pretraining corpora. Further details are provided in Appendix Section \ref{sec:srqa}.

\subsection{Dataset Headroom}
Many datasets used in prior work exhibit low headroom, meaning fine-tuning yields only marginal gains over the pretrained model. Table \ref{tab:headroom} quantifies headroom by comparing pretrained versus fine-tuned accuracy and NLL. We observe essentially no headroom on ARC-Challenge (with similar phenomena observed on ARC-Easy, OpenBookQA, and BoolQ as well), while Winogrande retains moderate headroom. In contrast, datasets newly introduced in BAG, such as Circuit Logic and SRQA, show large improvements after adaptation, providing a cleaner testbed for adaptation methods. Finally, MathVerse (and MMStar) remain challenging in a different way: performance is mediocre both before and after fine-tuning, consistent with a setting where the pretrained model lacks prior knowledge and the available training signal is weak.
\begin{figure*}
    \centering
    \includegraphics[width=\linewidth]{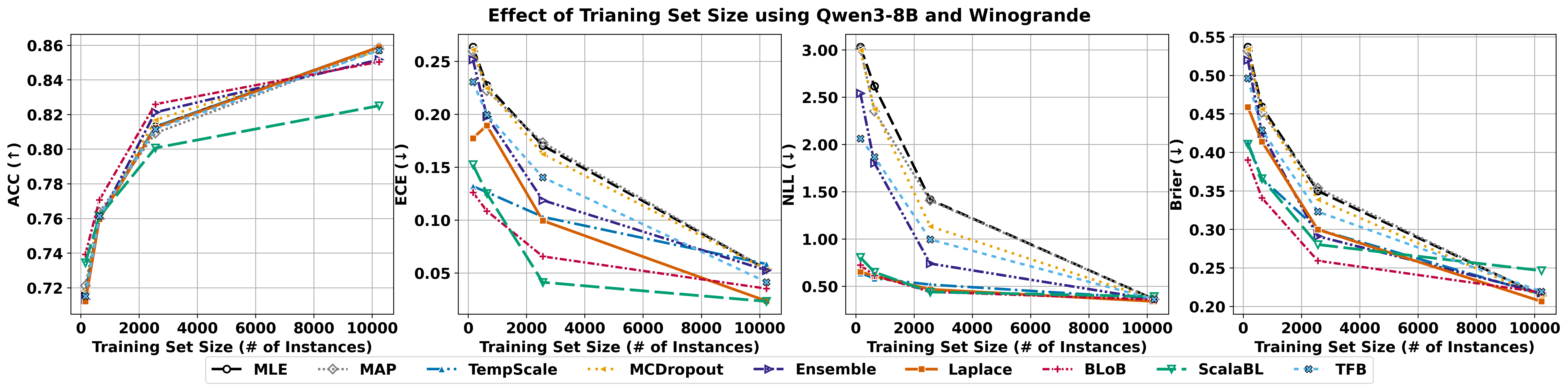}
    \caption{Effect of training set size using the Winograde dataset. We see that as the size of training set increases all methods converge to similar performance.}
    \label{fig:winogrande_size_comparsion}
\end{figure*}

\subsection{Implementation Details and Hyperparameters}
 BAG began as a fork of the \texttt{bayesian-peft} library introduced by \cite{blob}, which provided the implementations for BLoB and TFB. The original code for Laplace and ScalaBL was integrated directly into the framework. For all experiments, we use the recent Qwen3 \citep{qwen3} and Qwen3-VL \citep{qwen3vl} family of models. For the text-only models we consider sizes from 0.6 billion parameters to 14 billion base parameters and for the VLMs we use the instruction-tuned versions with 2, 4, and 8 billion base parameters. Following prior work, we apply the LoRA adapters to the query and value projection linear layers in each attention layer of the models as well as in the final classification output layer using a LoRA rank of $r=8$. Although beyond the scope of this paper, using BAG, is it straightforward to change the LoRA rank as well as perform efficient last layer Bayesianization \citep{harrison2024variational} by modifying the \texttt{hydra} configuration. Unless stated otherwise, we train all methods for 5000 steps using the AdamW optimizer using a batch size of 4. For all methods which require multiple samples, we use a Bayesian model average of size $N=10$. All approach-specific hyperparameters were copied over from the original implementation. For all experiments, we report the mean over 4 random seeds. Most experiments were performed on a single node with 8 RTX 6000 GPUs each with 48 GB of memory, with all experiments only using a single GPU at a time. For some experiments with large memory requirements we used a single A100 GPU with 80GB of memory.

\begin{figure}
    \centering
    \includegraphics[width=\linewidth]{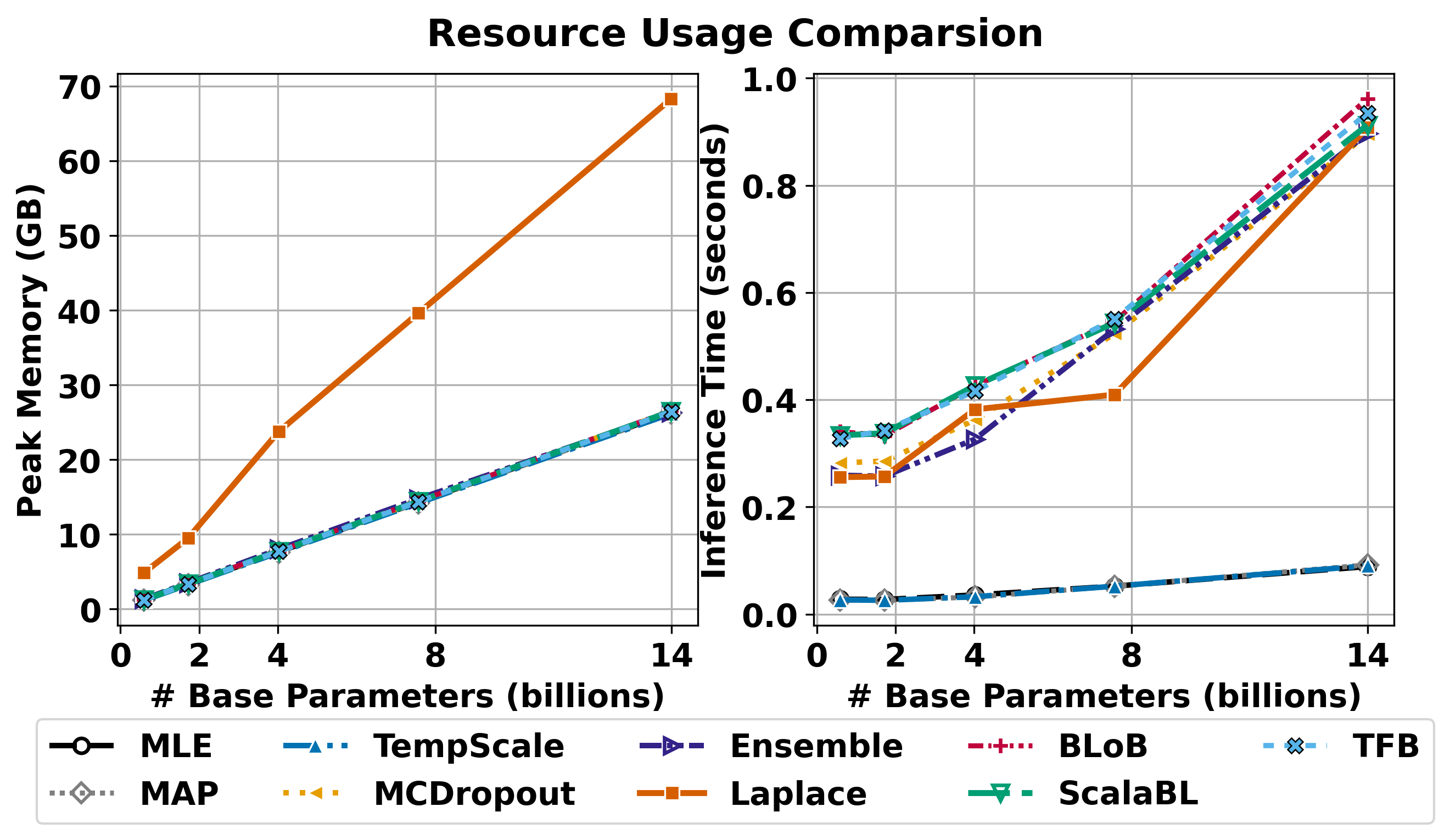}
    \caption{Inference time latency and peak memory usage for Qwen3 family using a batch size of 4 on Winogrande test set.}
    \label{fig:resource_use}
\end{figure}

\section{Experiments}
\subsection{In-distribution Calibration} \label{sec:exp_calibration}
We first discuss the evaluation protocol typically used in prior work. In Figure \ref{fig:obqa_id_calib} we show the in-distribution test performance for the Qwen3 model family. For visual clarity, we focus on the comparison between the recent state-of-the-art methods of BLoB and TFB against a simple MLE baseline, with and without temperature scaling. We first note that the accuracy of all approaches is within a tight bound of each other. This is due to the fact that the accuracy achievable on a dataset like OpenBookQA is highly determined by the model's base knowledge. We see that across all model sizes, the simple temperature-scaled MLE baseline has very competitive performance with BLoB and TFB across all UQ metrics. This highlights one of the fundamental shortcomings of the evaluation protocol of prior work, since this evaluation protocol is unable to distinguish between a complex Bayesian adaptation method and a simple baseline. We further notice that as the base model gets larger, the MLE gets more accurate and better calibrated. 

\begin{figure*}
    \centering
    \includegraphics[width=\linewidth]{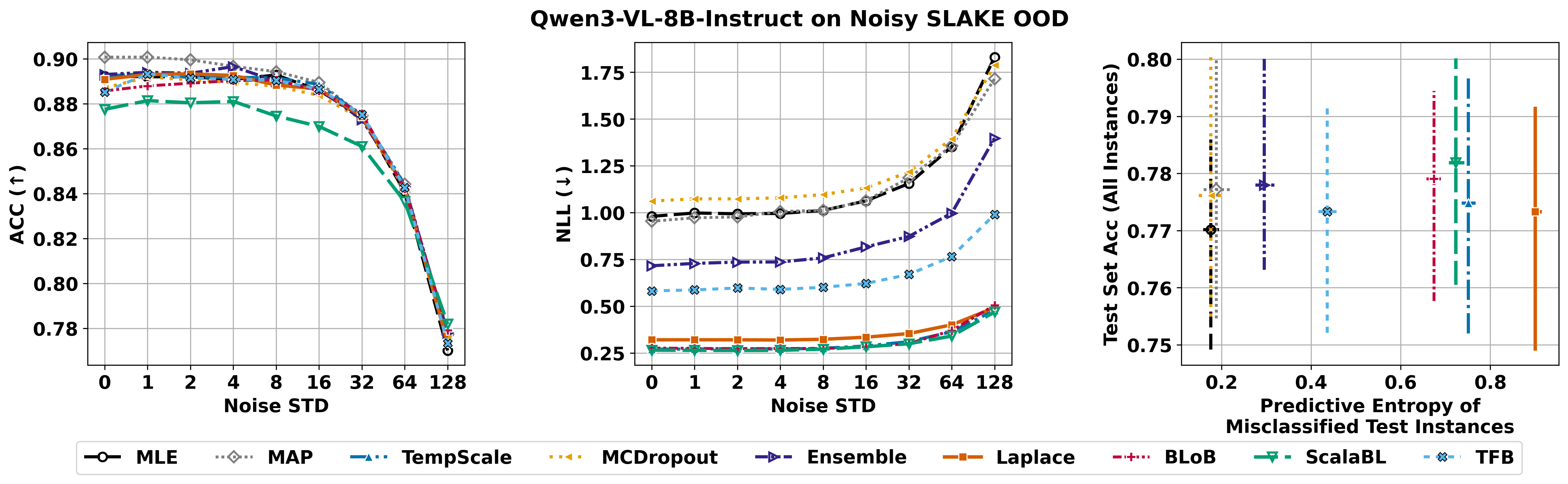}
    \caption{Evaluating approaches on SLAKE test images with Gaussian noise. Left and center plots show the degradation of performance in accuracy and NLL as noise increases, respectively. In the right most plot, we compare predictive entropy of \textit{misclassified} points against full test set accuracy using the highest level of noise ($\sigma=128$). Due to the noise in the evaluation process 10 evaluation runs we used for each of the 4 training seeds. Plots show the mean of those 40 points. Error bars show the IQR range.}
    \label{fig:noisy_slake_8B}
\end{figure*}
\subsection{Effect of Training Set Size}
Winogrande \citep{winogrande} is a binary commonsense reasoning dataset commonly used in prior work. It comes in a range of training set sizes from extra small (xs) to large (l), with prior work using small (s) and medium (m). These training sets are nested subsets of each other (e.g. all the instances in xs are in s, etc.), with a shared test set used for all sizes. In BAG, we support the full range of sizes and use it as a test of the effect of training set on the adaptation process. We adapt a Qwen3-8B model on each size of Winogrande and report results for all adaptation methods in Figure \ref{fig:winogrande_size_comparsion}. 

We observe the sensible behavior where the Bayesian techniques achieve better performance on the uncertainty metrics compared to an MLE or MAP point-estimate when the training set is small. However, as the training set grows, all methods converge to similar performance across metrics and methods, Bayesian or otherwise. This suggests that Bayesian adaptation methods are most useful in small training set regimes. 

\subsection{Resource Usage}
In Figure \ref{fig:resource_use} we provide the latency (seconds) and peak memory usage (GB) during model inference for a range of Qwen3 base models (0.6B to 14B parameters) using a batch size of 4 on the Winogrande test set. On the memory axis, we see that nearly all approaches have the same peak memory usage. This isn't surprising, as each sample from the posterior is computed sequentially, giving them all the same memory usage as a single MLE adapter. However, we can see the extreme memory demands of the Laplace approach, which scales poorly with base model size compared to all other methods.

In terms of latency, the approaches can be divided into two categories: single-shot approaches (MLE, MAP, and TempScale) and sampling-based approaches (all other approaches except Laplace). Naturally, these single-shot approaches, which need to compute only a single forward pass result in much lower latency. A major downside of these sampling-based approaches is that a forward pass of the entire network is computed for each sample ($N=10$ for all approaches) in order to compute the Bayesian model average. We see they are all much slower than MLE, but none is significantly faster than the others. We also see that even though Laplace is not a sampling-based approach, it still has worse latency than the MLE.

We further note that in general the training time of experiments using BAG is generally fast. A typical training run for a single seed takes approximately 5 to 20 minutes of wall-clock time, depending on the particular choice of method, model size, and dataset. In general, the low-rank weight updates of LoRA enable efficient training runs compared to the full parameter training of earlier Bayesian deep learning work.

\subsection{Robustness Experiments}
A key motivation of Bayesian approaches is to build models which more robustly respond to out-of-distribution (OOD) data encountered at test time. Prior work considered the OOD experiment of training on the OpenBookQA dataset and testing on either ARC-Easy/Challenge or MMLU. For similar reasons discussed above, we find this to be a poor test of the model's OOD performance, as most of these datasets have low headroom and the final performance is closely tied to the pretrained knowledge of the model. However, BAG still includes comparisons using prior datasets, where we produce a set of experiments using MMLU in Appendix Section \ref{sec:ood_mmlu}.

In contrast to prior work, in BAG we instead consider more meaningful OOD scenarios. We consider using the Circuit Logic dataset as a test of the model's ability to handle changes in input representation for the same underlying logical reasoning problem. Results for this set of experiments are shown in Appendix Section \ref{sec:ood_circuit}.

\begin{figure*}
    \centering
    \includegraphics[width=\linewidth]{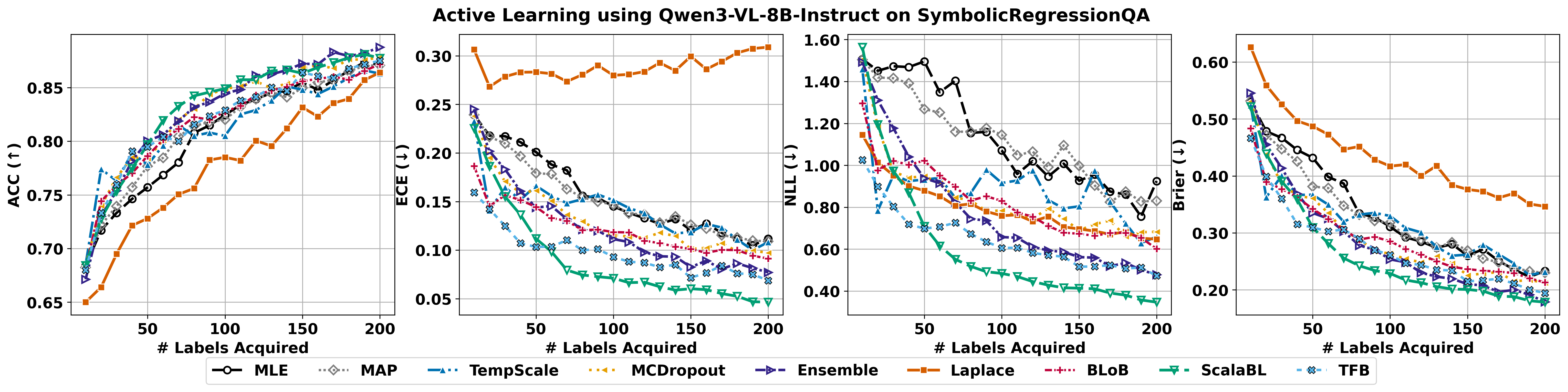}
    \caption{Active learning experiment on the SymbolicRegressionQA dataset using Qwen3-VL-8B-Instruct.}
    \label{fig:srqa_active_learn_8B}
\end{figure*}

\paragraph{SLAKE with Noise}
BAG extends to multi-modal datasets and enables out-of-distribution evaluations that are not possible in the text-only settings of prior work. In particular, we test robustness to test time visual corruption by adding Gaussian noise to the input image. We perform this experiment on the SLAKE dataset, since robustness to acquisition noise and artifacts especially important in medical settings. Concretely, we perturb grayscale pixel intensities with noise sampled from $\mathcal{N}(0,\sigma)$ (in $[0,255]$ pixel units) and evaluate performance across increasing noise levels $\sigma \in {[1,2,4,\ldots,128]}$ (see Appendix Figure \ref{fig:noise_slake}). In Figure~\ref{fig:noisy_slake_8B} we display results for Qwen3-VL-8B-Instruct on this experiment.

In the left hand and center figures we display how each method in BAG responds to increasing noise in terms of accuracy and NLL. As expected, across all approaches we see a strong decline in performance as the noise increases. In the right most figure we plot the average predictive entropy of misclassified test instances versus the full test set accuracy for the highest noise level ($\sigma = 128$). We would expect a Bayesian method to respond to noisy input by expressing higher uncertainty in its predictions. We see that approaches such as MLE and MAP maintain low entropy even in the face of high input noise, while Bayesian methods such as BLoB, ScalaBL, TFB, and Laplace have much higher uncertainty in this setting. However, we do again see competitive performance by the simple temperature scaling baseline.

\subsection{Active Learning}
Next we turn our attention to the problem of active learning. In this setting, we benchmark the performance of a model's predictive uncertainty by using it to guide the training data selection process and reduce the burden of labeling instances. We begin by taking the full training set to be the unlabeled data pool. We randomly select $k=10$ instances as the initial training set. Then, for $T$ iterations we perform the following loop: train an adaptation approach from scratch (i.e. cold start) using the current training set, use this trained model to compute an acquisition function $a$ for each instance in the pool, acquire the label for the top $k=10$ instances ranked by the acquisition function, and add them to the training set. For the sake of visualization, we also evaluate the model on a held aside test set after each iteration.

For non-sampling approaches such as MLE, MAP, and temperature scaling, we use the model's predictive entropy as the acquisition function. For the Bayesian adaptation approaches, we use Bayesian Active Learning by Disagreement (BALD) \citep{bald}. Further details on our active learning setup are discussed in Appendix Section \ref{sec:active_learning}. The results for this experiment using the SymbolicRegressionQA dataset are shown in Figure \ref{fig:srqa_active_learn_8B}.

We first note, in contrast to the evaluation setup of prior work this active learning setup shows a stronger and more definitive use case for Bayesian adaptation approaches. We see that in terms of NLL, all the Bayesian approaches (except Laplace) achieve stronger performance. We see especially strong performance using ScalaBL, which we hypothesize is due to its low-rank subspace providing a useful regularization when the training set is small.

An interesting feature of this active learning experiment is that it provides a different rank ordering of the Bayesian approaches compared to the one provided by the evaluation setup of prior work. In particular, we see that BLoB tends to perform comparably with MCDropout, a noticeable departure from the in-distribution calibration experiments of prior work. Beyond ScalaBL, we also see that Deep Ensembles and TFB achieve strong performance on this task. We also notice the worse performance of Laplace, which we suspect is caused by a low-quality Hessian approximation due to the small training set sizes used in active learning.

\section{Discussion}
Our experimentation via BAG has led to the following observations.
If the interest is primarily in having a well-calibrated classifier, then collecting a validation set and performing temperature scaling is likely more straightforward than using Bayesian adaptation. We find that Bayesian methods mostly shine when applied to domains where the base model has limited knowledge and the amount of fine-tuning dataset is low. We find Bayesian methods express greater uncertainty for out-of-distribution test cases than point-wise estimates. Our results on active learning suggest that Bayesian adaptation may be useful for tasks like data curation, active testing, and decision-problems more broadly. 

Overall, the approaches of BLoB, TFB, and ScalaBL remain the most promising methods. We find that samples from these approximations often outperform deep ensembles while requiring less training time. We find that the Laplace LoRA approach has inconsistent performance across datasets and is continually held back by its high memory demands, which reduces its scalability to larger models.

\section{Conclusion}
In this work we introduced Bayesian Adaptation Gym (BAG), a modular and extensible framework for the development and benchmarking of low-rank Bayesian adaptation of multi-modal language models. BAG includes implementations of traditional Bayesian techniques and state-of-the-art adaptation approaches. BAG also introduces a larger set of datasets and tasks to tackle an existing gap in the literature, where the prior work used data with little headroom for improvement, and did not provide enough distinguishing power to determine the utility of these approaches. We also include VLMs in the benchmark to evaluate Bayesian methods on multi-modal tasks for the first time. Finally, with the inclusion of the active learning task, which requires decision-making under uncertainty, we show that Bayesian adaptation approaches do provide benefits when data is sparse and better calibration is necessary. We open-source BAG and aim to include more approaches and tasks as this area of research continues to expand.

%\begin{acknowledgements} % will be removed in pdf for initial submission,
						 % (without ‘accepted’ option in \documentclass)
                         % so you can already fill it to test with the
                         % ‘accepted’ class option
 %   Briefly acknowledge people and organizations here.

  %  \emph{All} acknowledgements go in this section.
%\end{acknowledgements}

% References
\bibliography{sources}

%\newpage
\onecolumn
\title{Bayesian Adaptation Gym: A Benchmark for the Bayesian Low-Rank Adaptation of Multi-Modal Language Models\\(Supplementary Material)}
\maketitle
\appendix

\section{Further Method Details} \label{sec:method_details}
In this section we provide further details about each implemented method in BAG.

\subsection{MLE and MAP}
As discussed above, MLE and MAP represent a simple LoRA fine-tuning with and without weight decay, respectively. For the MAP approach, we use a weight decay of 0.01.

\subsection{Deep Ensembles}
For deep ensembles \citep{deepensembles}, we train a set of $N=10$ MLE adapters in parallel. At test time, we average the  softmax probabilities across the ensemble members.

\subsection{MC Dropout}
In the MCDropout approach of \cite{mcdropout} we train in a similar fashion to the MLE approach, but set the LoRA dropout rate to be 0.1. Then at test time we perform $N=10$ forward passes of the model with dropout turned on and compute the average softmax probabilities across these samples.

\subsection{Laplace LoRA}
The Laplace LoRA approach of \cite{lap} is a post-hoc Bayesian adaptation method which starts from a standard fine-tuning LoRA checkpoint \(\theta_{\mathrm{MLE}}\). An isotropic Gaussian prior \(p(\theta)=\mathcal{N}(0,\lambda^{-1}I)\) is placed over the LoRA parameters and the posterior is approximated by a Laplace Gaussian:
\[
p(\theta\mid \D)\approx \mathcal{N}(\theta_{\mathrm{MLE}},\Sigma),\qquad \Sigma=(F+\lambda I)^{-1},
\]
where $F$ is the Fisher curvature, in a KFAC Kronecker-factorized form. 

For a test input \(\x^{*}\), the model is linearized around \(\theta_{\mathrm{MLE}}\):
\[
f_\theta(\x^{*}) \approx f_{\theta_{\mathrm{MLE}}}(\x^{*}) + J_{x^{*}}(\theta-\theta_{\mathrm{MLE}}),
\]
which induces an approximate Gaussian distribution over logits
\[
f_\theta(\x^{*}) \sim \mathcal{N}\!\bigl(f_{\theta_{\mathrm{MLE}}}(\x^{*}),\ \Lambda\bigr),\qquad \Lambda = J_{\x^{*}}\Sigma J_{\x^{*}}^T.
\]
Predictive uncertainty is obtained via Bayesian model averaging by sampling \(\tilde f = f_{\theta_{\mathrm{MLE}}} + L\xi\) where $LL^T = \Lambda$ and $\xi \sim \mathcal{N}(0,I)$. 

\subsection{Stochastic Variational Approaches}
\subsubsection{BLoB}
The state-of-the-art approach of BLoB \citep{blob} is based on stochastic variational inference over the LoRA parameters $\A$. More specifically, they follow the Bayes by Backprop  approach introduced by \cite{bbb}. That is, they learn a variational approximation $q_{\theta}(\A)$ which is taken to be a Gaussian distribution with mean $\mathbf{A}_{\mu}$ and variance $\mathbf{A}_{\sigma}$. Using the reparameterization trick \citep{vae}, samples can be projected from this low rank distribution into the full weight space:
\begin{align}
    \W_t = \W_0 + \B (\A_{\mu}+\A_{\sigma} \cdot \eps_t)
\end{align}
where $\eps_t \sim \N(0,1)$.

This process remains fully differentiable, allowing all the parameters ($\B$, $\A_{\mu}$, and $\A_{\sigma}$) to be learned end-to-end using stochastic gradient methods. The training loss for these approaches is then the standard evidence lower bound (ELBO) \citep{elbo}.
\begin{align}
    \mathcal{L}_t &= \log P(\D_t|\W_t) - \beta D_{KL}( q_{\theta}(\A)|| P(\A))
\end{align}
where $P(\A)$ is a standard Gaussian prior. During training we use a single sample from $q_{\theta}(\A)$ and at test time we use the average predictive distribution across $N=10$ samples.

\subsubsection{ScalaBL}
The ScalaBL approach of \cite{scalabl} follows the stochastic variational inference approach of BLoB, but instead performs inference in a $r$-dimensional subspace (where $r$ is the LoRA rank). That is, we learn a variational approximation over an $r$-dimensional vector $\s$ as a diagonal Gaussian distribution:
\begin{align}
    q_{\btheta}(\s)=\N(\s|\s_{\mu}, \diag{\s_{\sigma}})
\end{align}
with mean and variance parameters $\btheta=[\s_{\mu},\s_{\sigma}]$.
Like BLoB the reparameterization trick is used to generate weight full weight samples:
\begin{align}
    \W_t = \W_0 + \B \diag{\s_{\mu} + \s_{\sigma} \cdot \eps_t} \A
    \label{eq:reparam}
\end{align}
where $\eps_t \sim \N(0,1)$. This is again optimized using the ELBO \citep{elbo}:
\begin{align}
    \mathcal{L}_t &= \log P(\D_t|\W_t) - \beta D_{KL}( q_{\theta}(\s)|| P(\s))
\end{align}
ScalaBL enjoys greater parameter efficiency compared to BLoB as it only needs to learn $r$ variance parameters compared to needing to learn the full $\A_{\sigma}$ matrix which is typically $r \times d$, where $d$ is the embedding dimension of the model. Like with BLoB, we use a single posterior sample during training and $N=10$ samples during test.

\subsubsection{Training-Free Bayesianization}
The Training-Free Bayesianization (TFB) approach of \cite{tfb} begins with a trained LoRA MLE adapter $\A$ and $\B$. Then a singular value decomposition (SVD) is performed on the learned $\B$ matrix:
\begin{align}
    \U \text{diag}(\s) \V^T = \B
\end{align}
 where $\s \in \R^r$ is the vector of singular values and $\U \in \R^{n \times r}, \V \in \R^{r \times r}$ are the left and right singular vectors, respectively. Using this they construct an equivalent LoRA adapter with $\A'=\V^T\A$ and $\B' = \U \text{diag}(\s)$. They then construct a Gaussian variational approximation with mean $\V^T\A$ and diagonal covariance matrix given by $\text{diag}(\sigma_q / \s)$ repeated $d$ times.

 Here $\sigma_q$ is a global variance parameter that is shared across all adapted layers. Rather than using stochastic training, a simple binary search is used. At every iteration the NLL on an ``anchor'' dataset is computed. The goal of the binary search is to find a maximal $\sigma_q$ that keeps the ratio between the original NLL (i.e. when $\sigma_q=0$) and current NLL within a tolerance threshold $\epsilon$. 

 We use the training set used for fine-tuning as the anchor dataset. Following \cite{tfb} we set the starting range of the binary search to $\sigma_q \in [0.001, 0.2]$, and use a ratio tolerance of $\epsilon=0.003$.

\subsection{Active Learning} \label{sec:active_learning}
\begin{algorithm}[t]
\caption{Active Learning Loop}
\label{alg:active_learning}
\begin{algorithmic}[1]
\Require Unlabeled pool $\mathcal{U}$
\Require batch size $k$, downsample size $M$, number of iterations $T$
\Require Model initialization distribution $\pi(\theta)$
\Require Acquisition function $a(\x;\theta)$, oracle $\textsc{OracleLabel}(\cdot)$
\State $\mathcal{L} \gets \textsc{RandomSample}\big(\mathcal{U}, k \big)$ 
\State $\mathcal{U} \gets \mathcal{U} \setminus \mathcal{L}$
\For{$t = 1$ \textbf{to} $T$}
    \State $\theta_0 \sim \pi(\theta)$ \Comment{Cold start parameters}
    \State $\theta^{*} \gets \textsc{Train}(\mathcal{L}, \theta_0)$
    \State $\mathcal{S} \gets \textsc{RandomSample}\big(\mathcal{U}, M \big)$ \Comment{Downsample the pool for acquisition computation}
    \State $\mathcal{Q} \gets \textsc{Topk}\big( a(\x, \theta^{*}) \textsc{ for } \x \in S \big)$ \Comment{Add $k$ instances with highest acquisition function}
    \State $\mathbf{y}_{\mathcal{Q}} \gets \textsc{OracleLabel}(\mathcal{Q})$
    \State $\mathcal{L} \gets \mathcal{L} \cup \{(\x,y): \x \in \mathcal{Q},\, y \in \mathbf{y}_{\mathcal{Q}}\}$
    \State $\mathcal{U} \gets \mathcal{U} \setminus \mathcal{Q}$
\EndFor
\State \Return $\mathcal{L}$
\end{algorithmic}
\end{algorithm}
In BAG we apply recent Bayesian adaptation approaches to the problem of active learning for the first time. The active learning loop that we used is  shown in Algorithm \ref{alg:active_learning}. For each training run within the loop, we train for 1000 steps and start from randomly initialized adaptation parameters each time (i.e. cold start). We find that the main bottleneck in this loop is computing the acquisition function $a$ on each element in the unlabeled pool, which in practice is the training set of one of the datasets in BAG. For this reason we randomly downsample $M=1000$ points at every iteration and select the top $k=10$ points only from that subset. 

For point-wise approaches, such as MLE or MAP, we use the predictive entropy of the softmax probabilities output by the model as the acquisition function. For the Bayesian approaches which compute a softmax probabilties for $N$ samples from the posterior, we use Bayesian Active Learning by Disagreement (BALD) \citep{bald}:
\begin{equation}
a_{\mathrm{BALD}}(\x, \theta)
\;=\;
H\!\big[p(y\mid \x,\mathcal{L})\big]
\;-\;
\mathbb{E}_{p(\theta\mid \mathcal{L})}\!\left[\,H\!\big[p(y\mid \x,\theta)\big]\,\right].
\end{equation}
BALD scores a candidate point $\x$ by the mutual information between its unknown label $y$ and the current model parameters $\theta$ given the current labeled data $\mathcal{L}$. It prefers points whose predictive distribution is globally uncertain (first term) and for which different posterior samples of $\theta$ disagree strongly (second term).

\subsection{Metrics}\label{sec:metrics}
In BAG, we report four common metrics for evaluating predictive performance and uncertainty quantification: accuracy, expected calibration error (ECE), negative log-likelihood (NLL), and Brier score \citep{brier}. For a test set $\{(\x_n, y_n)\}_{n=1}^N$ and a probabilistic classifier $P_{\btheta}(y\mid \x)$, these metrics are defined as follows:

\paragraph{Accuracy:}
Accuracy measures the fraction of correct predictions:
\begin{align}
\text{ACC} = \frac{1}{N}\sum_{n=1}^N \mathbf{1}\!\left[\arg\max_{c} P_{\btheta}(y=c\mid \x_n) = y_n\right].
\end{align}

\paragraph{Expected calibration error:}
ECE measures the mismatch between confidence and empirical accuracy by binning predictions according to their confidence $\hat{p}_n \coloneqq \max_c P_{\btheta}(y=c\mid \x_n)$. Let $B_k$ be the set of indices whose confidences fall into bin $k\in\{1,\dots,K\}$:
\begin{align}
\text{ECE} = \sum_{k=1}^{K} \frac{|B_k|}{N}\left| \text{ACC}(B_k) - \text{conf}(B_k) \right|,
\end{align}
where $\text{ACC}(B_k)$ is the accuracy of the model on the instances in the $k$th bin and
\begin{align}
    \text{conf}(B_k)=\frac{1}{|B_k|}\sum_{n\in B_k}\hat{p}_n
\end{align}
Following prior work \citep{lap,blob,scalabl}, we use $K=15$ bins in all experiments.

\paragraph{Negative log-likelihood:}
NLL is the average negative log-probability assigned to the true label:
\begin{align}
\text{NLL} = -\frac{1}{N}\sum_{n=1}^N \log P_{\btheta}(y_n \mid \x_n).
\end{align}

\paragraph{Brier score:}
The Brier score measures the mean squared error between the predicted probabilities and the one-hot label vector. Let $\mathbf{p}_{\btheta}(\x_n)\in[0,1]^C$ denote the predicted class-probability vector over $C$ classes and let $\mathbf{e}_{y_n}$ be the one-hot vector for label $y_n$:
\begin{align}
\text{Brier score} = \frac{1}{N}\sum_{n=1}^N \left\| \mathbf{p}_{\btheta}(\x_n) - \mathbf{e}_{y_n} \right\|_2^2.
\end{align}

\section{Additional Dataset Details}
\subsection{Winogrande}
Winogrande \citep{winogrande} is a dataset for evaluating commonsense reasoning and co-reference resolution via fill-in-the-blank sentence-completion problems. The training set of Winogrande is available in 6 sizes: extra-small (xs), small (s), medium (m), large (l), and extra-large (xl). These training sets are nested subsets of each other (e.g. all the instances in xs are in s). There is then one test set that is used for all sizes. The dataset has no prior defined validation split, which we require for the temperature scaling method. We build a shared validation split using the instances are that are unique to the xl training set.

\newtcolorbox{promptbox}{
  colback=black!2,
  colframe=black!20,
  boxrule=0.3pt,
  arc=2pt,
  left=6pt,right=6pt,top=6pt,bottom=6pt,
}

\subsubsection{Prompt Format by Example}
\begin{promptbox}
\begin{verbatim}
For the sentence given below, select the answer that best fills in the 
blank (_) from the given choices.

Ian volunteered to eat Dennis's menudo after already having a bowl because _ 
despised eating intestine.

Choices:
A) Ian
B) Dennis
\end{verbatim}
\end{promptbox}

\subsubsection{Dataset Statistics}

\begin{table}[h!]
\centering
\small
\begin{tabular}{c|c|c|c|c}
\hline
\multirow{2}{*}{\textbf{Size}} & \multicolumn{3}{c|}{\textbf{\# Instances}} & \multirow{2}{*}{\textbf{\# Classes}} \\
\cline{2-4}
 & \textbf{Train} & \textbf{Validation} & \textbf{Test} & \\
\hline
xs & 160   & \multirow{4}{*}{30164} & \multirow{4}{*}{1267} & \multirow{4}{*}{2} \\
s  & 640   &                        &                        &                    \\
m  & 2558  &                        &                        &                    \\
l  & 10234 &                        &                        &                    \\
\hline
\end{tabular}
\caption{Winogrande statistics}
\label{tab:winogrande_stats}
\end{table}

\subsection{ARC}
The ARC (AI2 Reasoning Challenge) dataset \citep{arc} is a multiple-choice science question dataset split into two difficulties: Easy and Challenge.

\subsubsection{Prompt Format by Example}
\begin{promptbox}
\begin{verbatim}
Answer the multiple choice question below. 
Output the letter of your choice only.

Which land form is the result of the constructive force of a glacier?
Choices:
A) valleys carved by a moving glacier
B) piles of rocks deposited by a melting glacier
C) grooves created in a granite surface by a glacier
D) bedrock hills roughened by the passing of a glacier
\end{verbatim}
\end{promptbox}

\subsubsection{Dataset Statistics}
\begin{table}[h!]
\centering
\small
\begin{tabular}{c|c|c|c|c}
\hline
\multirow{2}{*}{\textbf{Difficulty}} & \multicolumn{3}{c|}{\textbf{\# Instances}} & \multirow{2}{*}{\textbf{\# Classes}} \\
\cline{2-4}
 & \textbf{Train} & \textbf{Validation} & \textbf{Test} & \\
\hline
Easy & 2251   & 570 & 2376 & \multirow{2}{*}{4} \\
Challenge  & 1119   &     299             &    1172    &    \\
\hline
\end{tabular}
\caption{ARC statistics}
\label{tab:arc_stats}
\end{table}

\subsection{OpenBookQA}
The OpenBookQA (OBQA) dataset \citep{obqa} is a multiple-choice elementary school science question dataset.

\subsubsection{Prompt Format by Example}
\begin{promptbox}
\begin{verbatim}
Answer the multiple choice question below. 
Output the letter of your choice only.

The sun is responsible for
Choices:
A) puppies learning new tricks
B) children growing up and getting old
C) flowers wilting in a vase
D) plants sprouting, blooming and wilting
\end{verbatim}
\end{promptbox}

\subsubsection{Dataset Statistics}
\begin{table}[h!]
\centering
\small
\begin{tabular}{c|c|c|c}
\hline
\multicolumn{3}{c|}{\textbf{\# Instances}} & \multirow{2}{*}{\textbf{\# Classes}} \\
\cline{1-3}
\textbf{Train} & \textbf{Validation} & \textbf{Test} & \\
\hline
4957 & 500 & 500 & 4 \\
\hline
\end{tabular}
\caption{OpenBookQA statistics}
\label{tab:obqa_stats}
\end{table}

\subsection{BoolQ}
BoolQ \citep{boolq} is a yes/no question answering dataset built from  Google search queries paired with Wikipedia passages. Each example asks whether the passage entails the answer to the question, making it a benchmark for reading comprehension and natural language reasoning. In early experiments we found that BoolQ tended to have low headroom and was expensive to run due to the long text passages. For these reasons we do not report results on this dataset in this paper. However, the dataset is still fully supported in BAG.

\subsubsection{Prompt Format by Example}
\begin{promptbox}
\begin{verbatim}
Read the passage below and answer the question with the words 'true' or 'false'.

Passage: Windows Movie Maker (formerly known as Windows Live Movie Maker in 
Windows 7) is a discontinued video editing software by Microsoft. 
It is a part of Windows Essentials software suite and offers the ability to 
create and edit videos as well as to publish them on OneDrive, Facebook, 
Vimeo, YouTube, and Flickr.

Question: is windows movie maker part of windows essentials?
\end{verbatim}
\end{promptbox}

\subsubsection{Dataset Statistics}
\begin{table}[h!]
\centering
\small
\begin{tabular}{c|c|c|c}
\hline
\multicolumn{3}{c|}{\textbf{\# Instances}} & \multirow{2}{*}{\textbf{\# Classes}} \\
\cline{1-3}
\textbf{Train} & \textbf{Validation} & \textbf{Test} & \\
\hline
9427 & 0 & 3270 & 2 \\
\hline
\end{tabular}
\caption{BoolQ statistics}
\label{tab:boolq_stats}
\end{table}

\subsection{Circuit Logic} \label{sec:circuit_logic}
We adapt the Circuit Logic task from the recently purposed Reasoning Gym of \cite{reasoninggym}. The model must determine the output truth value of a randomly generated Boolean circuit displayed in unicode, given input assignments for each variable variables. Because each circuit can be represented equivalently as a logical formula, we have the ability to test a model's ability to generalize to different input representations for the same problem (see example below).

\subsubsection{Prompt Format by Example}
\begin{promptbox}
    \centering
    \includegraphics[width=0.3\linewidth]{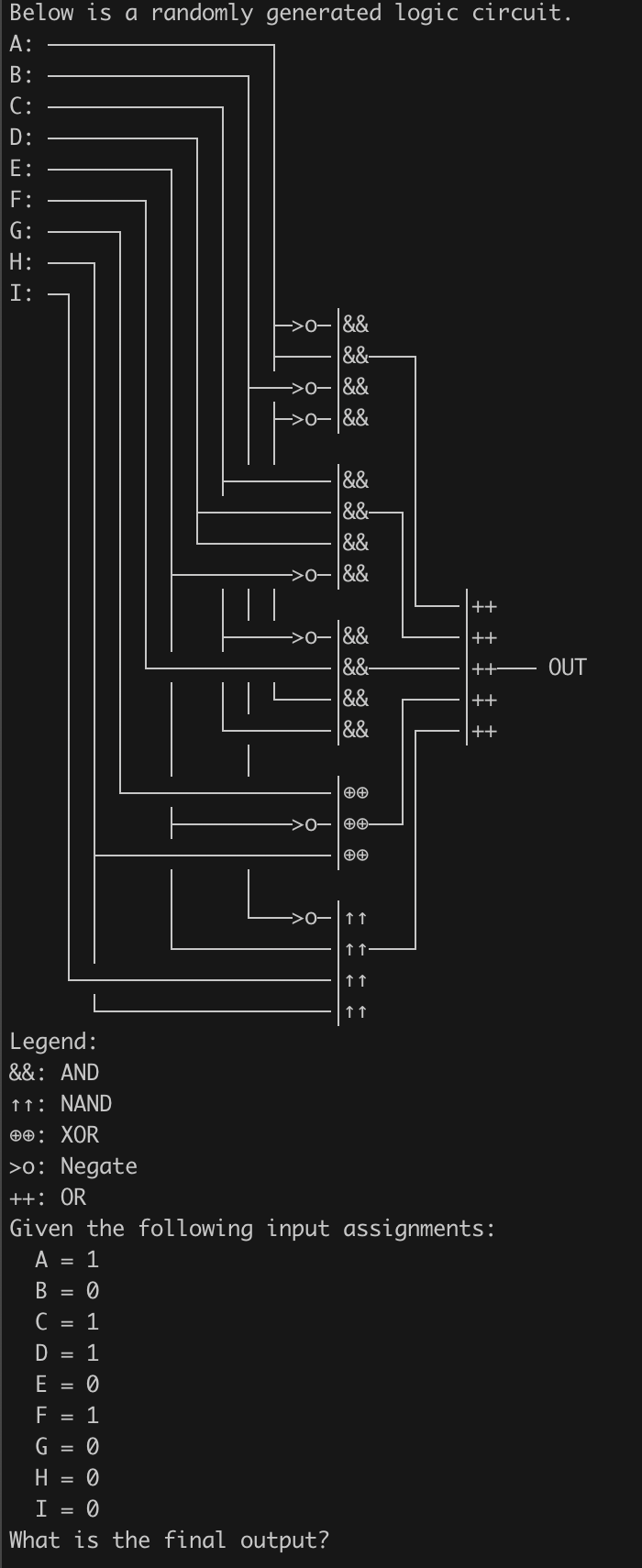}
    \includegraphics[width=0.5\linewidth]{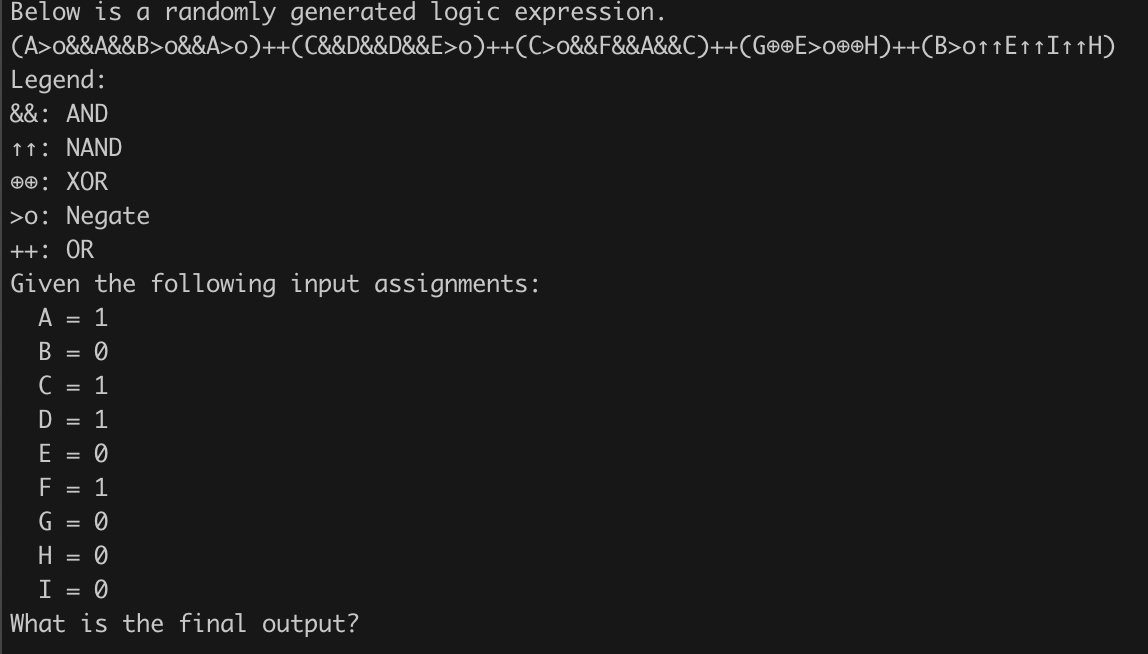}
\end{promptbox}

\subsubsection{Dataset Statistics}
\begin{table}[h!]
\centering
\small
\begin{tabular}{c|c|c|c}
\hline
\multicolumn{3}{c|}{\textbf{\# Instances}} & \multirow{2}{*}{\textbf{\# Classes}} \\
\cline{1-3}
\textbf{Train} & \textbf{Validation} & \textbf{Test} & \\
\hline
10000 & 1000 & 1000 & 2 \\
\hline
\end{tabular}
\caption{Circuit Logic statistics}
\label{tab:circuit_logic_stats}
\end{table}

\subsection{Slake} \label{sec:slake}
The SLAKE dataset of \cite{slake} tests a model's ability to understand radiology images in the medical domain. We use the subset of the full dataset which consists of Yes/No questions in English.

\subsubsection{Prompt Format by Example}
\begin{promptbox}
\centering
\includegraphics[width=0.5\linewidth]{figs/slake_example.png}
\begin{verbatim}
Answer the following question as Yes or No only.
Does the picture contain liver?
\end{verbatim}
\end{promptbox}

\subsubsection{Examples with Noise}
\begin{figure}[h!]
    \centering
    \includegraphics[width=0.98\linewidth]{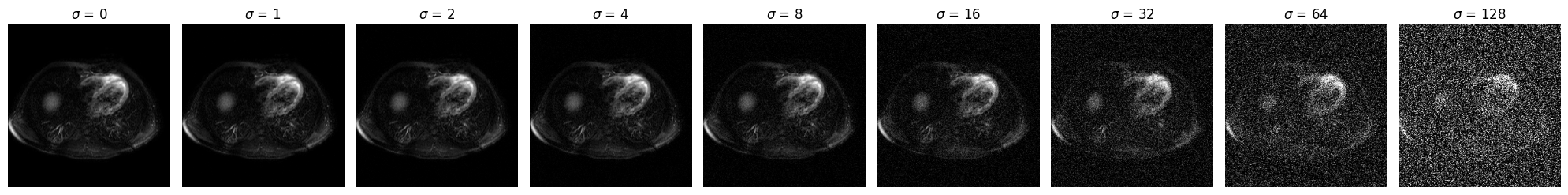}
    \caption{SLAKE image with increasing amounts of Gaussian pixel noise, $\sigma \in [0,1,2,4,\dots,128]$}.
    \label{fig:noise_slake}
\end{figure}

\subsubsection{Dataset Statistics}
\begin{table}[h!]
\centering
\small
\begin{tabular}{c|c|c|c}
\hline
\multicolumn{3}{c|}{\textbf{\# Instances}} & \multirow{2}{*}{\textbf{\# Classes}} \\
\cline{1-3}
\textbf{Train} & \textbf{Validation} & \textbf{Test} & \\
\hline
1681 & 355 & 358 & 2 \\
\hline
\end{tabular}
\caption{SLAKE statistics}
\label{tab:slake_stats}
\end{table}

\subsection{MathVerse} \label{sec:mathverse}
The MathVerse dataset of \cite{mathverse} tests a model's ability to answer mathematical reasoning problems which can only be answered correctly by understanding the associated input image. We randomly generate a train/validation/test split for use for this dataset.

\subsubsection{Prompt Format by Example}
\begin{promptbox}
\centering
\includegraphics[width=0.5\linewidth]{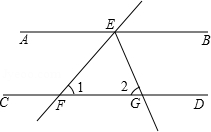}
\begin{verbatim}
Answer the multiple choice question below. 
Output the letter of your choice only.

As shown in the figure, angle 1 = 50.0, then angle 2 is equal to ()

Choices:
A:50°
B:60°
C:65°
D:90°
\end{verbatim}
\end{promptbox}

\subsubsection{Dataset Statistics}
\begin{table}[h!]
\centering
\small
\begin{tabular}{c|c|c|c}
\hline
\multicolumn{3}{c|}{\textbf{\# Instances}} & \multirow{2}{*}{\textbf{\# Classes}} \\
\cline{1-3}
\textbf{Train} & \textbf{Validation} & \textbf{Test} & \\
\hline
1465 & 315 & 315 & 4 \\
\hline
\end{tabular}
\caption{MathVerse statistics}
\label{tab:mathverse_stats}
\end{table}

\subsection{MMStar} \label{sec:mmstar}
The MMStar dataset of \cite{mmstar} tests a model's visual understanding and reasoning abilities using data cases where the model must understand the input image in order to correctly answer the question. We randomly generate a train/validation/test split for use for this dataset.

\subsubsection{Prompt Format by Example}
\begin{promptbox}
\centering
\includegraphics[width=0.5\linewidth]{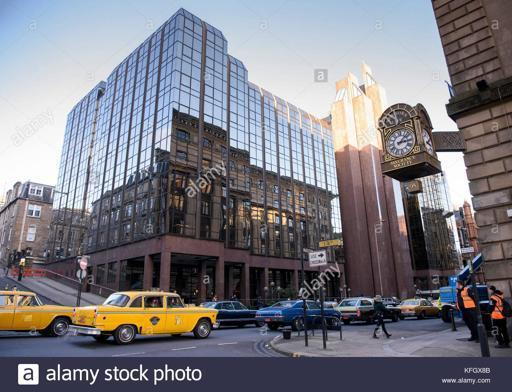}
\begin{verbatim}
Answer the multiple choice question below. 
Output the letter of your choice only.

What is the position of the blue car in the image?
Options: 
A: parked on the sidewalk
B: driving on the road
C: parked on the grass
D: parked on the road
\end{verbatim}
\end{promptbox}

\subsubsection{Dataset Statistics}
\begin{table}[h!]
\centering
\small
\begin{tabular}{c|c|c|c}
\hline
\multicolumn{3}{c|}{\textbf{\# Instances}} & \multirow{2}{*}{\textbf{\# Classes}} \\
\cline{1-3}
\textbf{Train} & \textbf{Validation} & \textbf{Test} & \\
\hline
1465 & 225 & 225 & 4 \\
\hline
\end{tabular}
\caption{MMStar statistics}
\label{tab:mmstar_stats}
\end{table}

\subsection{SymbolicRegressionQA} \label{sec:srqa}
In BAG we introduce a novel dataset that we call SymbolicRegressionQA (SRQA). Using the data generation code of \cite{snip}, we can generate a near finite supply of symbolic functions $f$ along with numerical data $(x,y)$ such that $x=f(x)$. We configure the data generator to generate only 1 dimensional functions so that we can render them as plots (see example below). We then sample 3 additional expressions to create a 4-way multiple choice question. The model is then tasked with picking which of the four options generated the plot.
\subsubsection{Prompt Format by Example}
\begin{promptbox}
\centering
\includegraphics[width=0.5\linewidth]{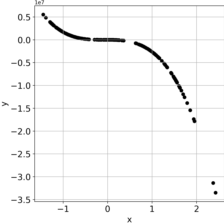}
\begin{verbatim}
For the provided image of a plot, which of following formulas best describes 
the relationship between the variables? 

Output the letter of your choice only.

Choices:
A) y = 5.75 - 0.85*x
B) y = -42.2124*exp(8.29*x) + 1.35*Abs(8.48*x - 0.693) - 0.697
C) y = -2137999.7563*(x + 0.0751)**3 + 62.6*sin(0.54*x + 0.072) + 0.524
D) y = 74.5*x - 1.68 + 39.2/(0.4401*(0.0785*x + 1)**2*cos(5.4*x - 0.163) - 5.86
\end{verbatim}
\end{promptbox}

\subsubsection{Dataset Statistics}
\begin{table}[h!]
\centering
\small
\begin{tabular}{c|c|c|c}
\hline
\multicolumn{3}{c|}{\textbf{\# Instances}} & \multirow{2}{*}{\textbf{\# Classes}} \\
\cline{1-3}
\textbf{Train} & \textbf{Validation} & \textbf{Test} & \\
\hline
10000 & 1000 & 1000 & 4 \\
\hline
\end{tabular}
\caption{SymbolicRegressionQA statistics}
\label{tab:srqa_stats}
\end{table}

\section{Additional Results}
In this section we provide a number of tables which display additional experiment results.

\subsection{In-distribution Results}
In the Tables \ref{tab:id_qwen06} through \ref{tab:id_qwen14} we display in-distribution results using the datasets of prior work: ARC-Easy, ARC-Challenge, and OpenBookQA (obqa) across all model sizes. This is followed by results on the Winogrande dataset for all sizes in Tables \ref{tab:id_qwen06_ws} through \ref{tab:id_qwen14_ws}. Then we show in-distribution performance for our image-based datasets: SLAKE, MMStar, MathVerse, and SymbolicRegressionQA (SRQA) in Tables \ref{tab:id_vlm_2b} through \ref{tab:id_vlm_8b}.

\subsubsection{OOD Results: OBQA -> MMLU}\label{sec:ood_mmlu}
We next consider out-of-distribution (ODD) experiments. We start with OOD experiments similar to prior work where we first train an adapter on the OpenBookQA dataset and then test on various topics from the MMLU dataset \citep{mmlu}. We note that in contrast to prior work we use the MMLU-Redux2.0 dataset which fixes a number of known issues in the original MMLU dataset \citep{mmluredux}. The results of these experiments are displayed in Tables \ref{tab:mmlu_ood_06B} through \ref{tab:mmlu_ood_14B}.

\subsubsection{OOD: Circuit Logic Representations} \label{sec:ood_circuit}
Next we present results on the out-of-distribution experiments using the Circuit Logic dataset. We train on a model using the circuit representation and then evaluate using a test of examples in the expression representation, and vice versa. The results for these experiments are displayed in  Tables \ref{tab:mmlu_clogic_06B} throuhg \ref{tab:mmlu_clogic_8B}. We see that across all approaches the model is able to generalize well from circuit representation to expression representation, but not the other way around. 

\subsection{Effect of Hyperparameters}
We next explore the effect of various hyperparameters. BAG's integration with \texttt{hydra} and the \texttt{ray} distributed computation runtime makes it simple to sweep over different hyperparameters just by modifying a bash script. Furthermore, its integration with the \texttt{ray} ecosystem  enables straightforward application of more advanced hyperparameter tuning algorithms, such as Bayesian optimization, which is an exciting direction for future work to select critical hyperparameters such as the LoRA rank $r$ or model size.

\subsection{Effect of LoRA rank $r$}
\begin{figure}[h!]
    \centering
    \includegraphics[width=\linewidth]{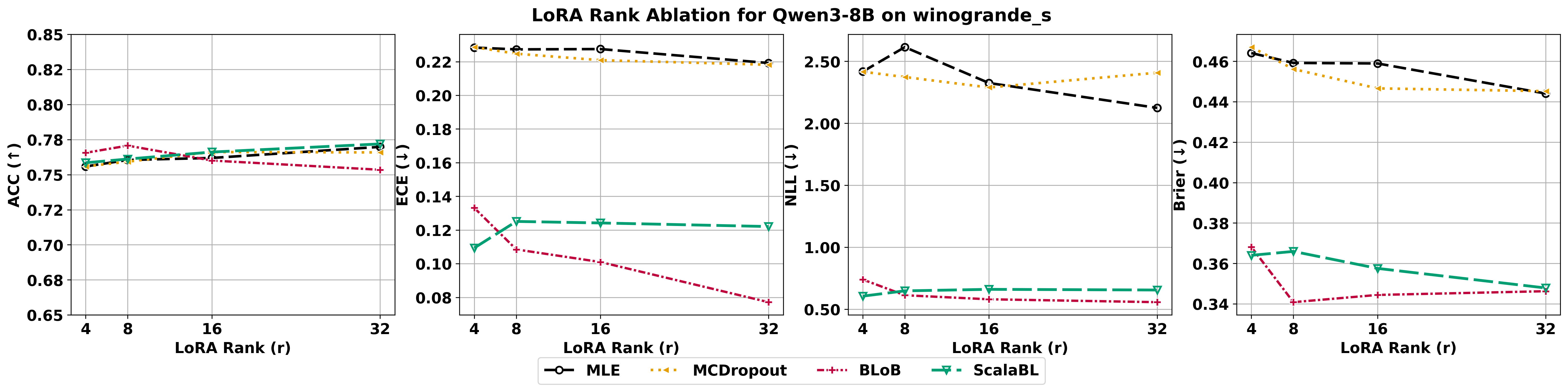}
    \caption{Ablation of LoRA rank on Winogrande-Small using Qwen3-8B.}
    \label{fig:lora_ablation_ws}
\end{figure}

\begin{figure}[h!]
    \centering
    \includegraphics[width=\linewidth]{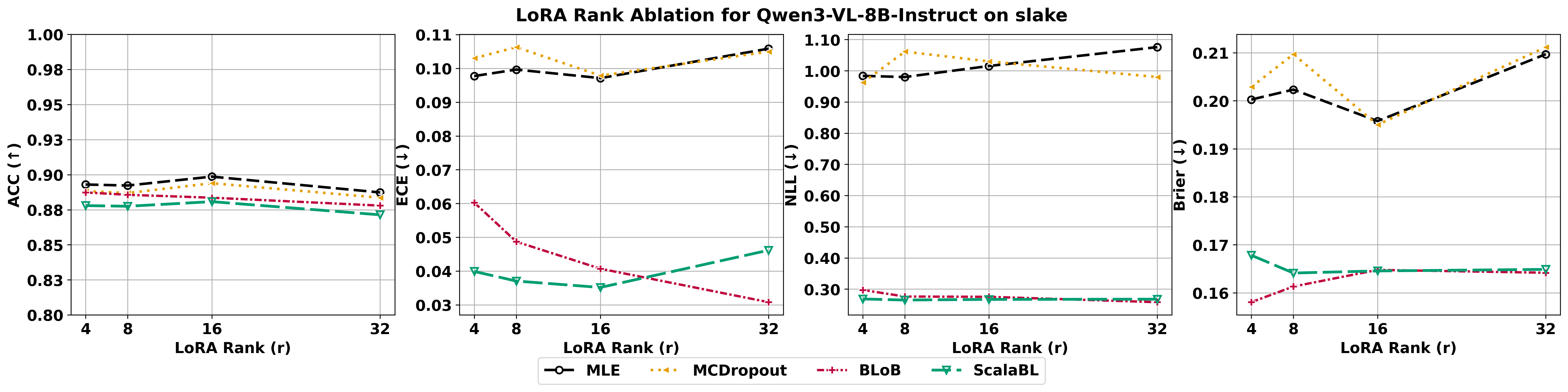}
    \caption{Ablation of LoRA rank on SLAKE using Qwen3-VL-8B-Instruct.}
    \label{fig:lora_ablation_slake}
\end{figure}

\begin{figure}[h!]
    \centering
    \includegraphics[width=\linewidth]{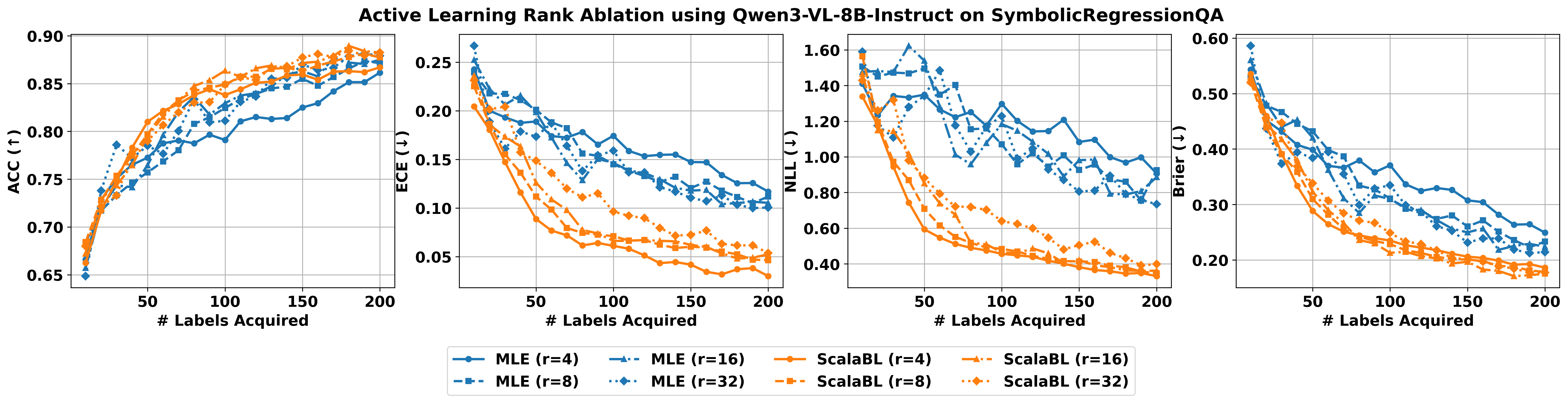}
    \caption{Ablation of LoRA rank on SRQA Active Learning using Qwen3-VL-8B-Instruct.}
    \label{fig:lora_ablation_active}
\end{figure}

\begin{figure}[h!]
    \centering
    \includegraphics[width=\linewidth]{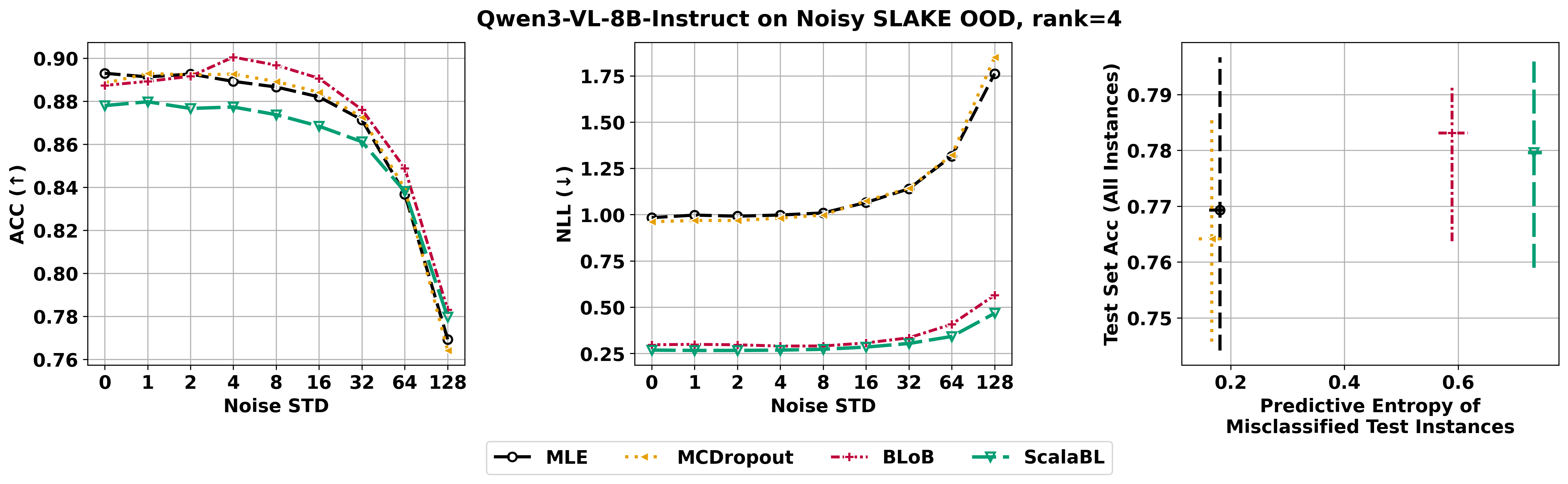}
    \caption{Effect of Noise on SLAKE dataset using Qwen3-VL-8B-Instruct with $r=4$}
    \label{fig:noisy_slake_r4}
\end{figure}

\begin{figure}[h!]
    \centering
    \includegraphics[width=\linewidth]{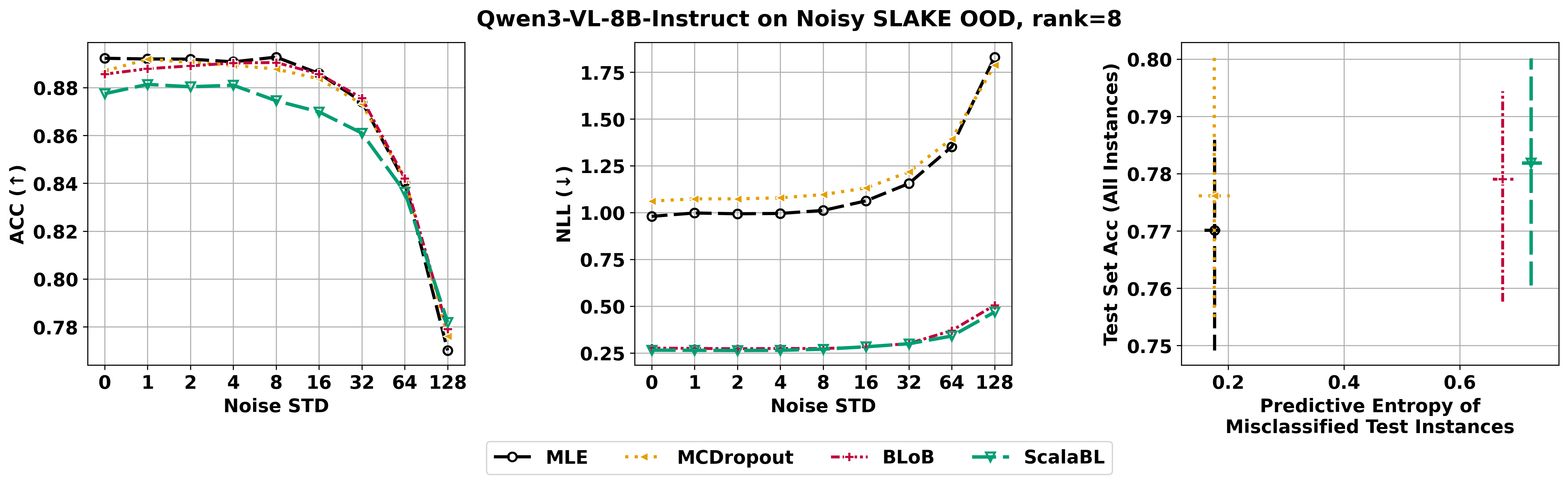}
    \caption{Effect of Noise on SLAKE dataset using Qwen3-VL-8B-Instruct with $r=8$}
    \label{fig:noisy_slake_r8}
\end{figure}

\begin{figure}[h!]
    \centering
    \includegraphics[width=\linewidth]{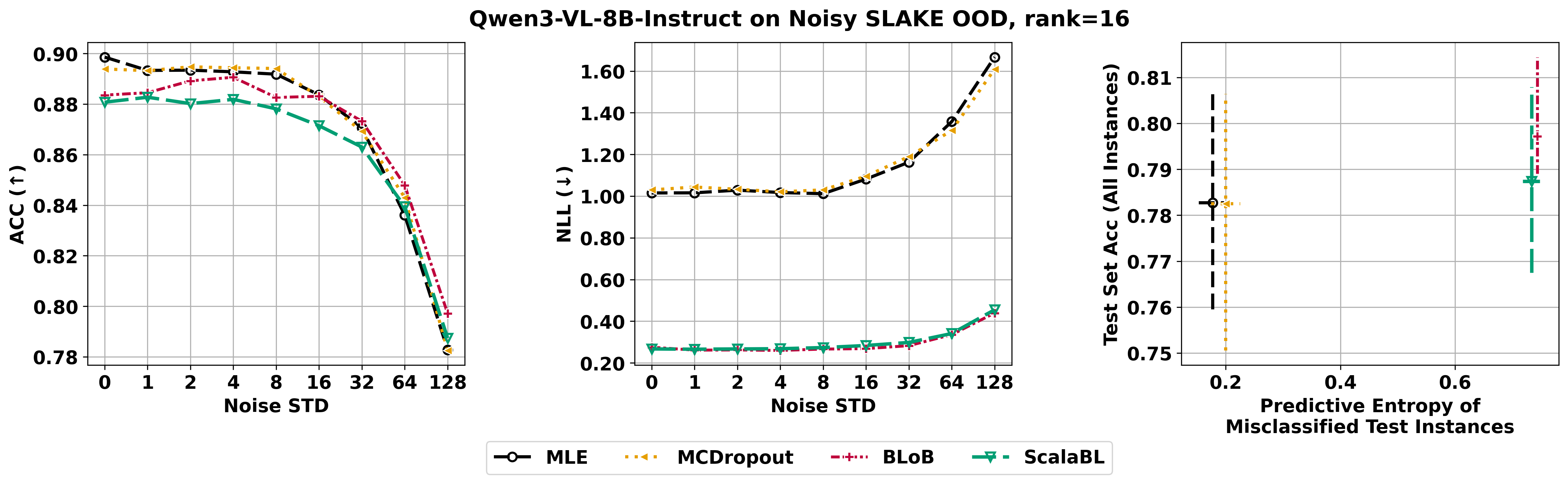}
    \caption{Effect of Noise on SLAKE dataset using Qwen3-VL-8B-Instruct with $r=16$}
    \label{fig:noisy_slake_r16}
\end{figure}

\begin{figure}[h!]
    \centering
    \includegraphics[width=\linewidth]{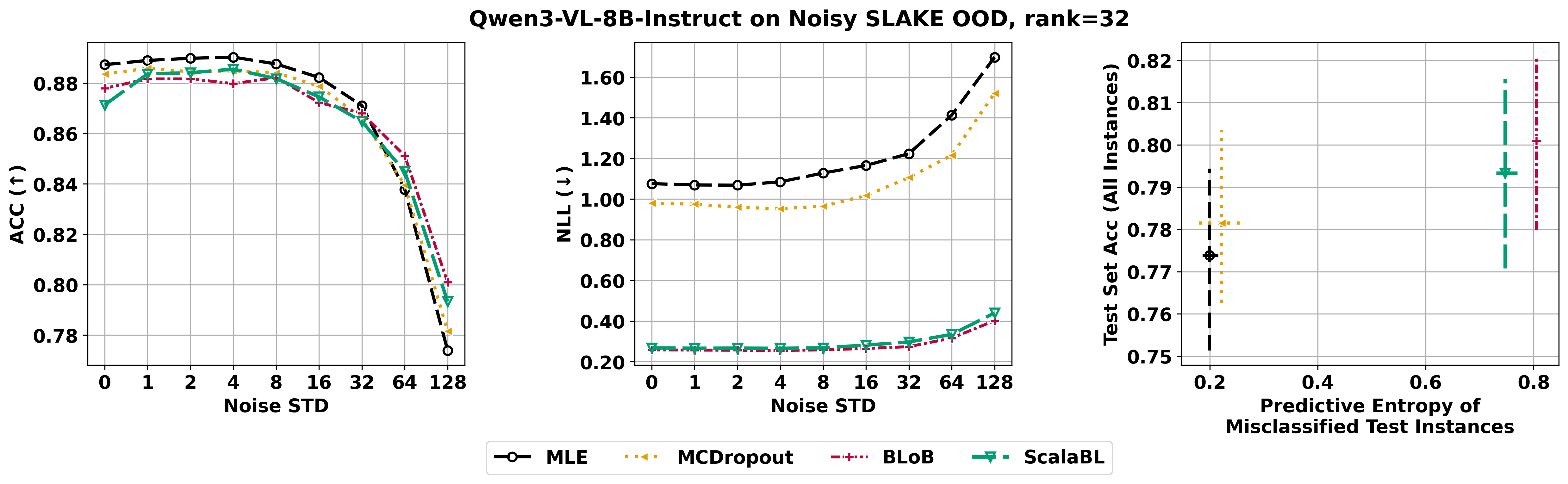}
    \caption{Effect of Noise on SLAKE dataset using Qwen3-VL-8B-Instruct with $r=32$}
    \label{fig:noisy_slake_r32}
\end{figure}

In Figures \ref{fig:lora_ablation_ws} through \ref{fig:noisy_slake_r32} the effect of the LoRA rank $r$ on the performance for a subset of the in-distrubtion, OOD, and active learning experiments. This is a key hyperparameter as it controls the general expressiveness of the fine-tuning process, as well as the dimensionality of the weight posterior for the Bayesian approaches. In general we notice the same high level trends regardless of rank. We see that the recent SoTA VI based approaches (BLoB and ScalaBL) strongly outperform simple baselines such as MLE or MCDropout. We find that the LoRA rank also has only a minor impact on the performance across in-distribution, OOD, and active learning results. Naturally, as the rank increases the capability of the fine-tuning process also increases. We suspect that for very large ranks this could lead to an overfitting effect. State of the art VI approaches can respond to this by increasing the weight of the KL term in the ELBO.

\subsection{Effect of Model Family}
\begin{figure}[h!]
    \centering
    \includegraphics[width=\linewidth]{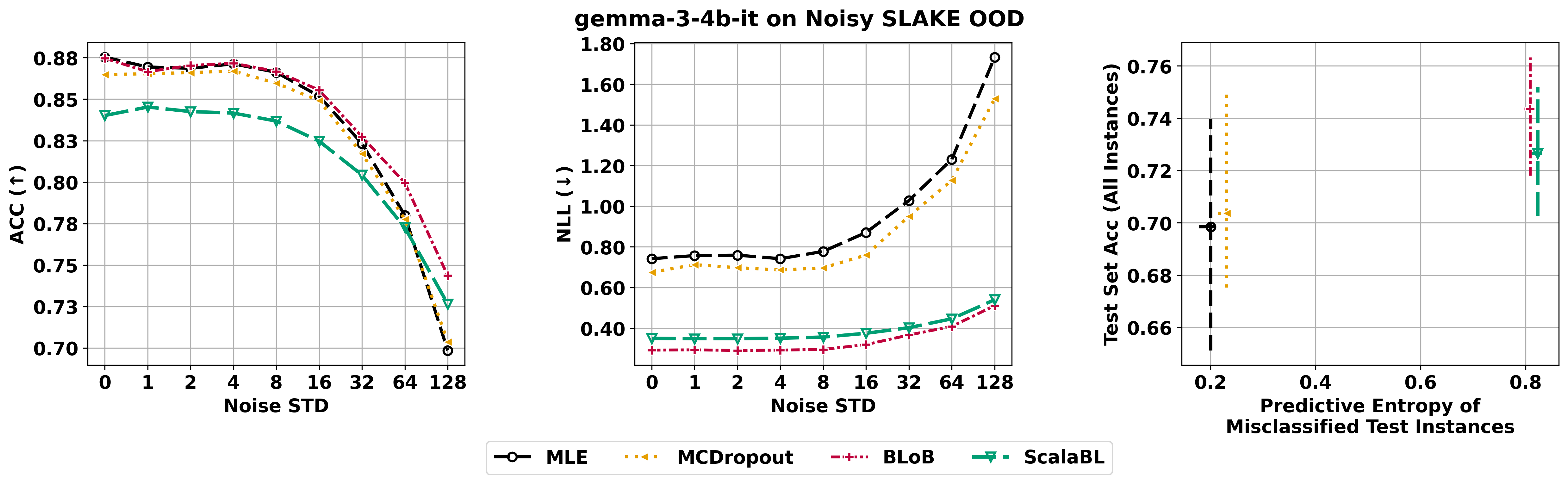}
    \caption{Effect of Noise on SLAKE dataset using Gemma-4B}
    \label{fig:noisy_slake_gemma}
\end{figure}

In this section we present results where compare the Qwen3 model used in all prior experiments with Google's Gemma3 VLM. In Tables \ref{tab:gemma_vs_qwen_id_ws} and \ref{tab:gemma_vs_qwen_id_slake} we display in-distribution results for the Winogrande Small and SLAKE datasets, respectively. In Figure \ref{fig:noisy_slake_gemma}, we present results OOD results using noisy SLAKE and Gemma3. In general we notice a similar outcome to the rank ablation above, with VI approaches out performing baselines. We see across most datasets that Gemma3-4B performs slightly worse than its Qwen3 analogue, which we attribute to Gemma being a weaker model overall compared to Qwen, even before fine-tuning.

\begin{table*}[h!]
\centering
\caption{In-Distribution Performance Comparison for Qwen3-0.6B}
\begin{tabular}{@{}ccccc@{}}
\toprule
\textbf{Metric} & \textbf{Method} & \textbf{ARC-Easy} & \textbf{ARC-Challenge} & \textbf{obqa} \\
\midrule
\multirow{9}{*}{\textbf{ACC ($\uparrow$)}}
& MLE & 0.807$_{\pm 0.004}$ & 0.601$_{\pm 0.004}$ & 0.695$_{\pm 0.004}$ \\
& MAP & 0.803$_{\pm 0.002}$ & 0.612$_{\pm 0.009}$ & 0.700$_{\pm 0.008}$ \\
& TempScale & 0.807$_{\pm 0.004}$ & 0.601$_{\pm 0.004}$ & 0.695$_{\pm 0.004}$ \\
& MCDropout & 0.806$_{\pm 0.005}$ & 0.608$_{\pm 0.009}$ & 0.700$_{\pm 0.009}$ \\
& Ensemble & 0.817$_{\pm 0.005}$ & 0.623$_{\pm 0.005}$ & 0.712$_{\pm 0.008}$ \\
& Laplace & 0.809$_{\pm 0.003}$ & 0.583$_{\pm 0.011}$ & 0.696$_{\pm 0.003}$ \\
& BLoB & 0.826$_{\pm 0.003}$ & 0.633$_{\pm 0.008}$ & 0.702$_{\pm 0.010}$ \\
\bottomrule
\end{tabular}
\label{tab:id_qwen06}
\end{table*}
\begin{table*}[h!]
\centering
\caption{In-Distribution Performance Comparison for Qwen3-1.7B}
\begin{tabular}{@{}ccccc@{}}
\toprule
\textbf{Metric} & \textbf{Method} & \textbf{ARC-Easy} & \textbf{ARC-Challenge} & \textbf{obqa} \\
\midrule
\multirow{9}{*}{\textbf{ACC ($\uparrow$)}}
& MLE & 0.905$_{\pm 0.003}$ & 0.780$_{\pm 0.005}$ & 0.819$_{\pm 0.012}$ \\
& MAP & 0.904$_{\pm 0.005}$ & 0.778$_{\pm 0.008}$ & 0.818$_{\pm 0.010}$ \\
& TempScale & 0.905$_{\pm 0.003}$ & 0.780$_{\pm 0.005}$ & 0.819$_{\pm 0.012}$ \\
& MCDropout & 0.907$_{\pm 0.004}$ & 0.778$_{\pm 0.007}$ & 0.818$_{\pm 0.009}$ \\
& Ensemble & 0.909$_{\pm 0.003}$ & 0.789$_{\pm 0.006}$ & 0.815$_{\pm 0.007}$ \\
& Laplace & 0.900$_{\pm 0.003}$ & 0.771$_{\pm 0.005}$ & 0.823$_{\pm 0.011}$ \\
& BLoB & 0.917$_{\pm 0.003}$ & 0.791$_{\pm 0.004}$ & 0.816$_{\pm 0.007}$ \\
\bottomrule
\end{tabular}
\label{tab:id_qwen17}
\end{table*}
\begin{table*}[h!]
\centering
\caption{In-Distribution Performance Comparison for Qwen3-4B}
\begin{tabular}{@{}ccccc@{}}
\toprule
\textbf{Metric} & \textbf{Method} & \textbf{ARC-Easy} & \textbf{ARC-Challenge} & \textbf{obqa} \\
\midrule
\multirow{9}{*}{\textbf{ACC ($\uparrow$)}}
& MLE & 0.954$_{\pm 0.003}$ & 0.887$_{\pm 0.002}$ & 0.912$_{\pm 0.014}$ \\
& MAP & 0.954$_{\pm 0.002}$ & 0.885$_{\pm 0.006}$ & 0.918$_{\pm 0.006}$ \\
& TempScale & 0.954$_{\pm 0.003}$ & 0.887$_{\pm 0.002}$ & 0.912$_{\pm 0.014}$ \\
& MCDropout & 0.953$_{\pm 0.002}$ & 0.885$_{\pm 0.002}$ & 0.916$_{\pm 0.007}$ \\
& Ensemble & 0.956$_{\pm 0.001}$ & 0.891$_{\pm 0.004}$ & 0.914$_{\pm 0.001}$ \\
& Laplace & 0.947$_{\pm 0.004}$ & 0.855$_{\pm 0.010}$ & 0.916$_{\pm 0.009}$ \\
& BLoB & 0.960$_{\pm 0.002}$ & 0.888$_{\pm 0.004}$ & 0.913$_{\pm 0.005}$ \\
\bottomrule
\end{tabular}
\label{tab:id_qwen4}
\end{table*}
\begin{table*}[h!]
\centering
\caption{In-Distribution Performance Comparison for Qwen3-8B}
\begin{tabular}{@{}ccccc@{}}
\toprule
\textbf{Metric} & \textbf{Method} & \textbf{ARC-Easy} & \textbf{ARC-Challenge} & \textbf{obqa} \\
\midrule
\multirow{9}{*}{\textbf{ACC ($\uparrow$)}}
& MLE & 0.966$_{\pm 0.002}$ & 0.912$_{\pm 0.003}$ & 0.924$_{\pm 0.008}$ \\
& MAP & 0.967$_{\pm 0.004}$ & 0.910$_{\pm 0.003}$ & 0.925$_{\pm 0.003}$ \\
& TempScale & 0.966$_{\pm 0.002}$ & 0.912$_{\pm 0.003}$ & 0.924$_{\pm 0.008}$ \\
& MCDropout & 0.965$_{\pm 0.002}$ & 0.913$_{\pm 0.003}$ & 0.924$_{\pm 0.002}$ \\
& Ensemble & 0.968$_{\pm 0.002}$ & 0.917$_{\pm 0.002}$ & 0.920$_{\pm 0.004}$ \\
& Laplace & 0.963$_{\pm 0.005}$ & 0.879$_{\pm 0.009}$ & 0.924$_{\pm 0.004}$ \\
& BLoB & 0.973$_{\pm 0.002}$ & 0.917$_{\pm 0.002}$ & 0.923$_{\pm 0.006}$ \\
\bottomrule
\end{tabular}
\label{tab:id_qwen8}
\end{table*}
\begin{table*}[h!]
\centering
\caption{In-Distribution Performance Comparison for Qwen3-14B}
\begin{tabular}{@{}ccccc@{}}
\toprule
\textbf{Metric} & \textbf{Method} & \textbf{ARC-Easy} & \textbf{ARC-Challenge} & \textbf{obqa} \\
\midrule
\multirow{9}{*}{\textbf{ACC ($\uparrow$)}}
& MLE & 0.980$_{\pm 0.001}$ & 0.935$_{\pm 0.002}$ & 0.946$_{\pm 0.005}$ \\
& MAP & 0.979$_{\pm 0.001}$ & 0.935$_{\pm 0.002}$ & 0.946$_{\pm 0.007}$ \\
& TempScale & 0.980$_{\pm 0.001}$ & 0.935$_{\pm 0.002}$ & 0.946$_{\pm 0.005}$ \\
& MCDropout & 0.980$_{\pm 0.001}$ & 0.935$_{\pm 0.002}$ & 0.948$_{\pm 0.002}$ \\
& Ensemble & 0.979$_{\pm 0.001}$ & 0.935$_{\pm 0.002}$ & 0.957$_{\pm 0.003}$ \\
& Laplace & 0.972$_{\pm 0.003}$ & 0.901$_{\pm 0.011}$ & 0.947$_{\pm 0.005}$ \\
& BLoB & 0.979$_{\pm 0.001}$ & 0.936$_{\pm 0.003}$ & 0.952$_{\pm 0.005}$ \\
\bottomrule
\end{tabular}
\label{tab:id_qwen14}
\end{table*}

\begin{table*}[h!]
\centering
\caption{In-Distribution Performance Comparison for Qwen3-0.6B}
\begin{tabular}{@{}cccccc@{}}
\toprule
\textbf{Metric} & \textbf{Method} & \textbf{winogrande\_xs} & \textbf{winogrande\_s} & \textbf{winogrande\_m} & \textbf{winogrande\_l} \\
\midrule
\multirow{9}{*}{\textbf{ACC ($\uparrow$)}}
& MLE & 0.555$_{\pm 0.016}$ & 0.574$_{\pm 0.011}$ & 0.576$_{\pm 0.002}$ & 0.588$_{\pm 0.008}$ \\
& MAP & 0.543$_{\pm 0.015}$ & 0.574$_{\pm 0.010}$ & 0.575$_{\pm 0.004}$ & 0.586$_{\pm 0.012}$ \\
& TempScale & 0.555$_{\pm 0.016}$ & 0.574$_{\pm 0.011}$ & 0.576$_{\pm 0.002}$ & 0.588$_{\pm 0.008}$ \\
& MCDropout & 0.555$_{\pm 0.008}$ & 0.575$_{\pm 0.007}$ & 0.576$_{\pm 0.004}$ & 0.590$_{\pm 0.004}$ \\
& Ensemble & 0.557$_{\pm 0.005}$ & 0.573$_{\pm 0.002}$ & 0.586$_{\pm 0.008}$ & 0.599$_{\pm 0.003}$ \\
& Laplace & 0.554$_{\pm 0.012}$ & 0.574$_{\pm 0.009}$ & 0.576$_{\pm 0.008}$ & 0.589$_{\pm 0.004}$ \\
& BLoB & 0.502$_{\pm 0.013}$ & 0.501$_{\pm 0.012}$ & 0.528$_{\pm 0.032}$ & 0.581$_{\pm 0.009}$ \\
\bottomrule
\end{tabular}
\label{tab:id_qwen06_ws}
\end{table*}

\begin{table*}[h!]
\centering
\caption{In-Distribution Performance Comparison for Qwen3-1.7B}
\begin{tabular}{@{}cccccc@{}}
\toprule
\textbf{Metric} & \textbf{Method} & \textbf{winogrande\_xs} & \textbf{winogrande\_s} & \textbf{winogrande\_m} & \textbf{winogrande\_l} \\
\midrule
\multirow{9}{*}{\textbf{ACC ($\uparrow$)}}
& MLE & 0.578$_{\pm 0.003}$ & 0.597$_{\pm 0.005}$ & 0.623$_{\pm 0.002}$ & 0.657$_{\pm 0.005}$ \\
& MAP & 0.588$_{\pm 0.009}$ & 0.602$_{\pm 0.008}$ & 0.623$_{\pm 0.006}$ & 0.657$_{\pm 0.005}$ \\
& TempScale & 0.578$_{\pm 0.003}$ & 0.597$_{\pm 0.005}$ & 0.623$_{\pm 0.002}$ & 0.657$_{\pm 0.005}$ \\
& MCDropout & 0.584$_{\pm 0.009}$ & 0.599$_{\pm 0.005}$ & 0.627$_{\pm 0.010}$ & 0.659$_{\pm 0.003}$ \\
& Ensemble & 0.585$_{\pm 0.006}$ & 0.604$_{\pm 0.010}$ & 0.630$_{\pm 0.005}$ & 0.659$_{\pm 0.007}$ \\
& Laplace & 0.580$_{\pm 0.004}$ & 0.597$_{\pm 0.007}$ & 0.622$_{\pm 0.003}$ & 0.660$_{\pm 0.005}$ \\
& BLoB & 0.524$_{\pm 0.050}$ & 0.534$_{\pm 0.047}$ & 0.570$_{\pm 0.046}$ & 0.640$_{\pm 0.009}$ \\
\bottomrule
\end{tabular}
\label{tab:id_qwen17_ws}
\end{table*}

\begin{table*}[h!]
\centering
\caption{In-Distribution Performance Comparison for Qwen3-4B}
\begin{tabular}{@{}cccccc@{}}
\toprule
\textbf{Metric} & \textbf{Method} & \textbf{winogrande\_xs} & \textbf{winogrande\_s} & \textbf{winogrande\_m} & \textbf{winogrande\_l} \\
\midrule
\multirow{9}{*}{\textbf{ACC ($\uparrow$)}}
& MLE & 0.685$_{\pm 0.004}$ & 0.711$_{\pm 0.004}$ & 0.756$_{\pm 0.009}$ & 0.807$_{\pm 0.006}$ \\
& MAP & 0.687$_{\pm 0.007}$ & 0.711$_{\pm 0.013}$ & 0.762$_{\pm 0.005}$ & 0.805$_{\pm 0.005}$ \\
& TempScale & 0.685$_{\pm 0.004}$ & 0.711$_{\pm 0.004}$ & 0.756$_{\pm 0.009}$ & 0.807$_{\pm 0.006}$ \\
& MCDropout & 0.685$_{\pm 0.008}$ & 0.709$_{\pm 0.004}$ & 0.760$_{\pm 0.008}$ & 0.809$_{\pm 0.003}$ \\
& Ensemble & 0.680$_{\pm 0.002}$ & 0.721$_{\pm 0.008}$ & 0.778$_{\pm 0.008}$ & 0.814$_{\pm 0.003}$ \\
& Laplace & 0.680$_{\pm 0.005}$ & 0.712$_{\pm 0.006}$ & 0.756$_{\pm 0.008}$ & 0.805$_{\pm 0.007}$ \\
& BLoB & 0.695$_{\pm 0.006}$ & 0.727$_{\pm 0.005}$ & 0.769$_{\pm 0.008}$ & 0.800$_{\pm 0.005}$ \\
\bottomrule
\end{tabular}
\label{tab:id_qwen4_ws}
\end{table*}

\begin{table*}[h!]
\centering
\caption{In-Distribution Performance Comparison for Qwen3-8B}
\begin{tabular}{@{}cccccc@{}}
\toprule
\textbf{Metric} & \textbf{Method} & \textbf{winogrande\_xs} & \textbf{winogrande\_s} & \textbf{winogrande\_m} & \textbf{winogrande\_l} \\
\midrule
\multirow{9}{*}{\textbf{ACC ($\uparrow$)}}
& MLE & 0.717$_{\pm 0.007}$ & 0.760$_{\pm 0.005}$ & 0.813$_{\pm 0.004}$ & 0.859$_{\pm 0.004}$ \\
& MAP & 0.721$_{\pm 0.005}$ & 0.764$_{\pm 0.006}$ & 0.809$_{\pm 0.002}$ & 0.858$_{\pm 0.004}$ \\
& TempScale & 0.717$_{\pm 0.007}$ & 0.760$_{\pm 0.005}$ & 0.813$_{\pm 0.004}$ & 0.859$_{\pm 0.004}$ \\
& MCDropout & 0.718$_{\pm 0.004}$ & 0.759$_{\pm 0.003}$ & 0.817$_{\pm 0.002}$ & 0.857$_{\pm 0.004}$ \\
& Ensemble & 0.713$_{\pm 0.001}$ & 0.760$_{\pm 0.006}$ & 0.821$_{\pm 0.006}$ & 0.852$_{\pm 0.007}$ \\
& Laplace & 0.712$_{\pm 0.006}$ & 0.760$_{\pm 0.004}$ & 0.812$_{\pm 0.005}$ & 0.859$_{\pm 0.004}$ \\
& BLoB & 0.739$_{\pm 0.006}$ & 0.771$_{\pm 0.010}$ & 0.826$_{\pm 0.004}$ & 0.850$_{\pm 0.009}$ \\
\bottomrule
\end{tabular}
\label{tab:id_qwen8_ws}
\end{table*}

\begin{table*}[h!]
\centering
\caption{In-Distribution Performance Comparison for Qwen3-14B}
\begin{tabular}{@{}cccccc@{}}
\toprule
\textbf{Metric} & \textbf{Method} & \textbf{winogrande\_xs} & \textbf{winogrande\_s} & \textbf{winogrande\_m} & \textbf{winogrande\_l} \\
\midrule
\multirow{9}{*}{\textbf{ACC ($\uparrow$)}}
& MLE & 0.741$_{\pm 0.002}$ & 0.786$_{\pm 0.004}$ & 0.847$_{\pm 0.007}$ & 0.885$_{\pm 0.005}$ \\
& MAP & 0.744$_{\pm 0.006}$ & 0.787$_{\pm 0.006}$ & 0.848$_{\pm 0.006}$ & 0.884$_{\pm 0.001}$ \\
& TempScale & 0.741$_{\pm 0.002}$ & 0.786$_{\pm 0.004}$ & 0.847$_{\pm 0.007}$ & 0.885$_{\pm 0.005}$ \\
& MCDropout & 0.746$_{\pm 0.006}$ & 0.785$_{\pm 0.009}$ & 0.846$_{\pm 0.004}$ & 0.884$_{\pm 0.002}$ \\
& Ensemble & 0.748$_{\pm 0.007}$ & 0.794$_{\pm 0.003}$ & 0.856$_{\pm 0.006}$ & 0.867$_{\pm 0.004}$ \\
& Laplace & 0.741$_{\pm 0.004}$ & 0.785$_{\pm 0.004}$ & 0.846$_{\pm 0.007}$ & 0.887$_{\pm 0.003}$ \\
& BLoB & 0.766$_{\pm 0.007}$ & 0.799$_{\pm 0.008}$ & 0.847$_{\pm 0.006}$ & 0.883$_{\pm 0.003}$ \\
\bottomrule
\end{tabular}
\label{tab:id_qwen14_ws}
\end{table*}

\begin{table*}[h!]
\centering
\caption{In-Distribution Performance Comparison for Qwen3-VL-2B-Instruct}
\begin{tabular}{@{}cccccc@{}}
\toprule
\textbf{Metric} & \textbf{Method} & \textbf{slake} & \textbf{mmstar} & \textbf{MathVerse} & \textbf{srqa} \\
\midrule
\multirow{9}{*}{\textbf{ACC ($\uparrow$)}}
& MLE & 0.873$_{\pm 0.004}$ & 0.558$_{\pm 0.037}$ & 0.383$_{\pm 0.026}$ & 0.949$_{\pm 0.003}$ \\
& MAP & 0.874$_{\pm 0.001}$ & 0.560$_{\pm 0.022}$ & 0.395$_{\pm 0.024}$ & 0.949$_{\pm 0.002}$ \\
& TempScale & 0.873$_{\pm 0.004}$ & 0.561$_{\pm 0.033}$ & 0.380$_{\pm 0.022}$ & 0.949$_{\pm 0.003}$ \\
& MCDropout & 0.878$_{\pm 0.007}$ & 0.575$_{\pm 0.016}$ & 0.383$_{\pm 0.033}$ & 0.947$_{\pm 0.003}$ \\
& Ensemble & 0.876$_{\pm 0.009}$ & 0.586$_{\pm 0.012}$ & 0.417$_{\pm 0.004}$ & 0.925$_{\pm 0.004}$ \\
& Laplace & 0.873$_{\pm 0.002}$ & 0.562$_{\pm 0.013}$ & 0.379$_{\pm 0.043}$ & 0.948$_{\pm 0.004}$ \\
& BLoB & 0.869$_{\pm 0.006}$ & 0.540$_{\pm 0.015}$ & 0.425$_{\pm 0.027}$ & 0.942$_{\pm 0.003}$ \\
\bottomrule
\end{tabular}
\label{tab:id_vlm_2b}
\end{table*}

\begin{table*}[h!]
\centering
\caption{In-Distribution Performance Comparison for Qwen3-VL-4B-Instruct}
\begin{tabular}{@{}cccccc@{}}
\toprule
\textbf{Metric} & \textbf{Method} & \textbf{slake} & \textbf{mmstar} & \textbf{MathVerse} & \textbf{srqa} \\
\midrule
\multirow{9}{*}{\textbf{ACC ($\uparrow$)}}
& MLE & 0.894$_{\pm 0.005}$ & 0.609$_{\pm 0.021}$ & 0.512$_{\pm 0.026}$ & 0.962$_{\pm 0.003}$ \\
& MAP & 0.892$_{\pm 0.008}$ & 0.611$_{\pm 0.028}$ & 0.492$_{\pm 0.009}$ & 0.963$_{\pm 0.004}$ \\
& TempScale & 0.894$_{\pm 0.005}$ & 0.608$_{\pm 0.015}$ & 0.516$_{\pm 0.028}$ & 0.962$_{\pm 0.003}$ \\
& MCDropout & 0.899$_{\pm 0.004}$ & 0.630$_{\pm 0.023}$ & 0.518$_{\pm 0.017}$ & 0.964$_{\pm 0.003}$ \\
& Ensemble & 0.899$_{\pm 0.011}$ & 0.621$_{\pm 0.016}$ & 0.548$_{\pm 0.017}$ & 0.948$_{\pm 0.004}$ \\
& Laplace & 0.894$_{\pm 0.005}$ & 0.580$_{\pm 0.006}$ & 0.532$_{\pm 0.036}$ & 0.962$_{\pm 0.005}$ \\
& BLoB & 0.900$_{\pm 0.006}$ & 0.608$_{\pm 0.018}$ & 0.521$_{\pm 0.023}$ & 0.953$_{\pm 0.005}$ \\
\bottomrule
\end{tabular}
\label{tab:id_vlm_4b}
\end{table*}

\begin{table*}[h!]
\centering
\caption{In-Distribution Performance Comparison for Qwen3-VL-8B-Instruct}
\begin{tabular}{@{}cccccc@{}}
\toprule
\textbf{Metric} & \textbf{Method} & \textbf{slake} & \textbf{mmstar} & \textbf{MathVerse} & \textbf{srqa} \\
\midrule
\multirow{9}{*}{\textbf{ACC ($\uparrow$)}}
& MLE & 0.892$_{\pm 0.008}$ & 0.656$_{\pm 0.020}$ & 0.513$_{\pm 0.031}$ & 0.966$_{\pm 0.003}$ \\
& MAP & 0.901$_{\pm 0.006}$ & 0.632$_{\pm 0.035}$ & 0.527$_{\pm 0.031}$ & 0.964$_{\pm 0.005}$ \\
& TempScale & 0.892$_{\pm 0.008}$ & 0.653$_{\pm 0.025}$ & 0.511$_{\pm 0.030}$ & 0.965$_{\pm 0.002}$ \\
& MCDropout & 0.887$_{\pm 0.011}$ & 0.644$_{\pm 0.011}$ & 0.532$_{\pm 0.016}$ & 0.961$_{\pm 0.000}$ \\
& Ensemble & 0.893$_{\pm 0.005}$ & 0.653$_{\pm 0.016}$ & 0.571$_{\pm 0.018}$ & 0.952$_{\pm 0.003}$ \\
& Laplace & 0.891$_{\pm 0.009}$ & 0.610$_{\pm 0.025}$ & 0.512$_{\pm 0.025}$ & 0.965$_{\pm 0.002}$ \\
& BLoB & 0.886$_{\pm 0.008}$ & 0.638$_{\pm 0.017}$ & 0.569$_{\pm 0.025}$ & 0.962$_{\pm 0.004}$ \\
\bottomrule
\end{tabular}
\label{tab:id_vlm_8b}
\end{table*}

\begin{table*}[h!]
\centering
\caption{OOD Performance Comparison for Qwen3-0.6B}
\begin{tabular}{@{}cccccccc@{}}
\toprule
\textbf{Metric} & \textbf{Method} & \multicolumn{6}{c}{\textbf{Train on obqa}} \\
\cmidrule(lr){3-8}
 &  & \textbf{In-dist} & \multicolumn{5}{c}{\textbf{OOD (test set)}} \\
 &  &  & \textbf{MMLU-Bio} & \textbf{MMLU-CS} & \textbf{MMLU-Chem} & \textbf{MMLU-Math} & \textbf{MMLU-Physics} \\
\midrule
\multirow{9}{*}{\textbf{ACC ($\uparrow$)}} & MLE & 0.695$_{\pm 0.004}$ & 0.487$_{\pm 0.037}$ & 0.423$_{\pm 0.025}$ & 0.460$_{\pm 0.026}$ & 0.348$_{\pm 0.017}$ & 0.272$_{\pm 0.035}$ \\
 & MAP & 0.700$_{\pm 0.008}$ & 0.505$_{\pm 0.027}$ & 0.420$_{\pm 0.026}$ & 0.443$_{\pm 0.048}$ & 0.351$_{\pm 0.035}$ & 0.288$_{\pm 0.039}$ \\
 & TempScale & 0.695$_{\pm 0.004}$ & 0.487$_{\pm 0.037}$ & 0.423$_{\pm 0.025}$ & 0.460$_{\pm 0.026}$ & 0.348$_{\pm 0.017}$ & 0.272$_{\pm 0.035}$ \\
 & MCDropout & 0.700$_{\pm 0.009}$ & 0.513$_{\pm 0.030}$ & 0.412$_{\pm 0.000}$ & 0.460$_{\pm 0.046}$ & 0.356$_{\pm 0.046}$ & 0.280$_{\pm 0.039}$ \\
 & Ensemble & 0.712$_{\pm 0.008}$ & 0.526$_{\pm 0.018}$ & 0.436$_{\pm 0.020}$ & 0.470$_{\pm 0.023}$ & 0.384$_{\pm 0.014}$ & 0.275$_{\pm 0.026}$ \\
\bottomrule
\end{tabular}
\label{tab:mmlu_ood_06B}
\end{table*}

\begin{table*}[h!]
\centering
\caption{OOD Performance Comparison for Qwen3-1.7B}
\begin{tabular}{@{}cccccccc@{}}
\toprule
\textbf{Metric} & \textbf{Method} & \multicolumn{6}{c}{\textbf{Train on obqa}} \\
\cmidrule(lr){3-8}
 &  & \textbf{In-dist} & \multicolumn{5}{c}{\textbf{OOD (test set)}} \\
 &  &  & \textbf{MMLU-Bio} & \textbf{MMLU-CS} & \textbf{MMLU-Chem} & \textbf{MMLU-Math} & \textbf{MMLU-Physics} \\
\midrule
\multirow{9}{*}{\textbf{ACC ($\uparrow$)}} & MLE & 0.819$_{\pm 0.012}$ & 0.564$_{\pm 0.010}$ & 0.505$_{\pm 0.012}$ & 0.527$_{\pm 0.017}$ & 0.404$_{\pm 0.030}$ & 0.453$_{\pm 0.048}$ \\
 & MAP & 0.818$_{\pm 0.010}$ & 0.582$_{\pm 0.022}$ & 0.508$_{\pm 0.015}$ & 0.533$_{\pm 0.029}$ & 0.394$_{\pm 0.030}$ & 0.447$_{\pm 0.036}$ \\
 & TempScale & 0.819$_{\pm 0.012}$ & 0.564$_{\pm 0.010}$ & 0.505$_{\pm 0.012}$ & 0.527$_{\pm 0.017}$ & 0.404$_{\pm 0.030}$ & 0.453$_{\pm 0.048}$ \\
 & MCDropout & 0.818$_{\pm 0.009}$ & 0.574$_{\pm 0.021}$ & 0.490$_{\pm 0.018}$ & 0.540$_{\pm 0.023}$ & 0.386$_{\pm 0.021}$ & 0.472$_{\pm 0.022}$ \\
 & Ensemble & 0.815$_{\pm 0.007}$ & 0.566$_{\pm 0.018}$ & 0.505$_{\pm 0.012}$ & 0.537$_{\pm 0.033}$ & 0.364$_{\pm 0.034}$ & 0.470$_{\pm 0.036}$ \\
\bottomrule
\end{tabular}
\label{tab:mmlu_ood_17B}
\end{table*}

\begin{table*}[h!]
\centering
\caption{OOD Performance Comparison for Qwen3-4B}
\begin{tabular}{@{}cccccccc@{}}
\toprule
\textbf{Metric} & \textbf{Method} & \multicolumn{6}{c}{\textbf{Train on obqa}} \\
\cmidrule(lr){3-8}
 &  & \textbf{In-dist} & \multicolumn{5}{c}{\textbf{OOD (test set)}} \\
 &  &  & \textbf{MMLU-Bio} & \textbf{MMLU-CS} & \textbf{MMLU-Chem} & \textbf{MMLU-Math} & \textbf{MMLU-Physics} \\
\midrule
\multirow{9}{*}{\textbf{ACC ($\uparrow$)}} & MLE & 0.912$_{\pm 0.014}$ & 0.842$_{\pm 0.026}$ & 0.647$_{\pm 0.015}$ & 0.663$_{\pm 0.030}$ & 0.515$_{\pm 0.036}$ & 0.597$_{\pm 0.017}$ \\
 & MAP & 0.918$_{\pm 0.006}$ & 0.865$_{\pm 0.015}$ & 0.637$_{\pm 0.023}$ & 0.640$_{\pm 0.029}$ & 0.503$_{\pm 0.036}$ & 0.610$_{\pm 0.008}$ \\
 & TempScale & 0.912$_{\pm 0.014}$ & 0.842$_{\pm 0.026}$ & 0.647$_{\pm 0.015}$ & 0.663$_{\pm 0.030}$ & 0.515$_{\pm 0.036}$ & 0.597$_{\pm 0.017}$ \\
 & MCDropout & 0.916$_{\pm 0.007}$ & 0.862$_{\pm 0.018}$ & 0.634$_{\pm 0.013}$ & 0.653$_{\pm 0.036}$ & 0.513$_{\pm 0.022}$ & 0.620$_{\pm 0.022}$ \\
 & Ensemble & 0.914$_{\pm 0.001}$ & 0.857$_{\pm 0.008}$ & 0.629$_{\pm 0.008}$ & 0.680$_{\pm 0.019}$ & 0.497$_{\pm 0.028}$ & 0.598$_{\pm 0.013}$ \\
\bottomrule
\end{tabular}
\label{tab:mmlu_ood_4B}
\end{table*}

\begin{table*}[h!]
\centering
\caption{OOD Performance Comparison for Qwen3-8B}
\begin{tabular}{@{}cccccccc@{}}
\toprule
\textbf{Metric} & \textbf{Method} & \multicolumn{6}{c}{\textbf{Train on obqa}} \\
\cmidrule(lr){3-8}
 &  & \textbf{In-dist} & \multicolumn{5}{c}{\textbf{OOD (test set)}} \\
 &  &  & \textbf{MMLU-Bio} & \textbf{MMLU-CS} & \textbf{MMLU-Chem} & \textbf{MMLU-Math} & \textbf{MMLU-Physics} \\
\midrule
\multirow{9}{*}{\textbf{ACC ($\uparrow$)}} & MLE & 0.924$_{\pm 0.008}$ & 0.834$_{\pm 0.021}$ & 0.678$_{\pm 0.023}$ & 0.713$_{\pm 0.008}$ & 0.581$_{\pm 0.017}$ & 0.573$_{\pm 0.021}$ \\
 & MAP & 0.925$_{\pm 0.003}$ & 0.832$_{\pm 0.031}$ & 0.673$_{\pm 0.023}$ & 0.720$_{\pm 0.029}$ & 0.568$_{\pm 0.010}$ & 0.575$_{\pm 0.025}$ \\
 & TempScale & 0.924$_{\pm 0.008}$ & 0.834$_{\pm 0.021}$ & 0.678$_{\pm 0.023}$ & 0.713$_{\pm 0.008}$ & 0.581$_{\pm 0.017}$ & 0.573$_{\pm 0.021}$ \\
 & MCDropout & 0.924$_{\pm 0.002}$ & 0.849$_{\pm 0.013}$ & 0.675$_{\pm 0.013}$ & 0.713$_{\pm 0.013}$ & 0.598$_{\pm 0.028}$ & 0.557$_{\pm 0.029}$ \\
 & Ensemble & 0.920$_{\pm 0.004}$ & 0.849$_{\pm 0.015}$ & 0.688$_{\pm 0.018}$ & 0.717$_{\pm 0.013}$ & 0.581$_{\pm 0.024}$ & 0.575$_{\pm 0.030}$ \\
\bottomrule
\end{tabular}
\label{tab:mmlu_ood_8B}
\end{table*}

\begin{table*}[h!]
\centering
\caption{OOD Performance Comparison for Qwen3-14B}
\begin{tabular}{@{}cccccccc@{}}
\toprule
\textbf{Metric} & \textbf{Method} & \multicolumn{6}{c}{\textbf{Train on obqa}} \\
\cmidrule(lr){3-8}
 &  & \textbf{In-dist} & \multicolumn{5}{c}{\textbf{OOD (test set)}} \\
 &  &  & \textbf{MMLU-Bio} & \textbf{MMLU-CS} & \textbf{MMLU-Chem} & \textbf{MMLU-Math} & \textbf{MMLU-Physics} \\
\midrule
\multirow{9}{*}{\textbf{ACC ($\uparrow$)}} & MLE & 0.946$_{\pm 0.005}$ & 0.878$_{\pm 0.012}$ & 0.740$_{\pm 0.018}$ & 0.687$_{\pm 0.013}$ & 0.682$_{\pm 0.024}$ & 0.665$_{\pm 0.013}$ \\
 & MAP & 0.946$_{\pm 0.007}$ & 0.878$_{\pm 0.014}$ & 0.747$_{\pm 0.020}$ & 0.693$_{\pm 0.022}$ & 0.689$_{\pm 0.024}$ & 0.680$_{\pm 0.008}$ \\
 & TempScale & 0.946$_{\pm 0.005}$ & 0.878$_{\pm 0.012}$ & 0.740$_{\pm 0.018}$ & 0.687$_{\pm 0.013}$ & 0.682$_{\pm 0.024}$ & 0.665$_{\pm 0.013}$ \\
 & MCDropout & 0.948$_{\pm 0.002}$ & 0.885$_{\pm 0.010}$ & 0.745$_{\pm 0.024}$ & 0.693$_{\pm 0.011}$ & 0.694$_{\pm 0.019}$ & 0.677$_{\pm 0.022}$ \\
 & Ensemble & 0.957$_{\pm 0.003}$ & 0.857$_{\pm 0.016}$ & 0.711$_{\pm 0.023}$ & 0.680$_{\pm 0.012}$ & 0.667$_{\pm 0.018}$ & 0.650$_{\pm 0.017}$ \\
\bottomrule
\end{tabular}
\label{tab:mmlu_ood_14B}
\end{table*}

\begin{table*}[h!]
\centering
\caption{OOD Performance Comparison for Qwen3-0.6B}
\begin{tabular}{@{}cccccc@{}}
\toprule
\textbf{Metric} & \textbf{Method} & \multicolumn{2}{c}{\textbf{Train on circuit\_logic}} & \multicolumn{2}{c}{\textbf{Train on expression\_logic}} \\
\cmidrule(lr){3-4} \cmidrule(lr){5-6}
 &  & \textbf{In-dist} & \textbf{expression\_logic (OOD)} & \textbf{In-dist} & \textbf{circuit\_logic (OOD)} \\
\midrule
\multirow{9}{*}{\textbf{ACC ($\uparrow$)}} & MLE & 0.841$_{\pm 0.001}$ & 0.831$_{\pm 0.004}$ & 0.848$_{\pm 0.011}$ & 0.625$_{\pm 0.018}$ \\
 & MAP & 0.843$_{\pm 0.005}$ & 0.832$_{\pm 0.004}$ & 0.847$_{\pm 0.006}$ & 0.623$_{\pm 0.015}$ \\
 & TempScale & 0.841$_{\pm 0.001}$ & 0.831$_{\pm 0.004}$ & 0.848$_{\pm 0.011}$ & 0.625$_{\pm 0.018}$ \\
 & MCDropout & 0.843$_{\pm 0.008}$ & 0.831$_{\pm 0.005}$ & 0.845$_{\pm 0.011}$ & 0.628$_{\pm 0.025}$ \\
 & Ensemble & 0.845$_{\pm 0.003}$ & 0.830$_{\pm 0.003}$ & 0.848$_{\pm 0.008}$ & 0.619$_{\pm 0.003}$ \\
 & Laplace & 0.840$_{\pm 0.007}$ & 0.830$_{\pm 0.003}$ & 0.845$_{\pm 0.008}$ & 0.626$_{\pm 0.016}$ \\
\bottomrule
\end{tabular}
\label{tab:mmlu_clogic_06B}
\end{table*}

\begin{table*}[h!]
\centering
\caption{OOD Performance Comparison for Qwen3-1.7B}
\begin{tabular}{@{}cccccc@{}}
\toprule
\textbf{Metric} & \textbf{Method} & \multicolumn{2}{c}{\textbf{Train on circuit\_logic}} & \multicolumn{2}{c}{\textbf{Train on expression\_logic}} \\
\cmidrule(lr){3-4} \cmidrule(lr){5-6}
 &  & \textbf{In-dist} & \textbf{expression\_logic (OOD)} & \textbf{In-dist} & \textbf{circuit\_logic (OOD)} \\
\midrule
\multirow{9}{*}{\textbf{ACC ($\uparrow$)}} & MLE & 0.838$_{\pm 0.004}$ & 0.784$_{\pm 0.038}$ & 0.846$_{\pm 0.002}$ & 0.606$_{\pm 0.007}$ \\
 & MAP & 0.848$_{\pm 0.006}$ & 0.746$_{\pm 0.099}$ & 0.846$_{\pm 0.004}$ & 0.608$_{\pm 0.011}$ \\
 & TempScale & 0.838$_{\pm 0.004}$ & 0.784$_{\pm 0.038}$ & 0.846$_{\pm 0.002}$ & 0.606$_{\pm 0.007}$ \\
 & MCDropout & 0.842$_{\pm 0.010}$ & 0.791$_{\pm 0.037}$ & 0.847$_{\pm 0.006}$ & 0.613$_{\pm 0.009}$ \\
 & Ensemble & 0.829$_{\pm 0.003}$ & 0.758$_{\pm 0.093}$ & 0.842$_{\pm 0.005}$ & 0.610$_{\pm 0.004}$ \\
 & Laplace & 0.841$_{\pm 0.005}$ & 0.783$_{\pm 0.042}$ & 0.846$_{\pm 0.004}$ & 0.606$_{\pm 0.006}$ \\
\bottomrule
\end{tabular}
\label{tab:mmlu_clogic_17B}
\end{table*}

\begin{table*}[h!]
\centering
\caption{OOD Performance Comparison for Qwen3-4B}
\begin{tabular}{@{}cccccc@{}}
\toprule
\textbf{Metric} & \textbf{Method} & \multicolumn{2}{c}{\textbf{Train on circuit\_logic}} & \multicolumn{2}{c}{\textbf{Train on expression\_logic}} \\
\cmidrule(lr){3-4} \cmidrule(lr){5-6}
 &  & \textbf{In-dist} & \textbf{expression\_logic (OOD)} & \textbf{In-dist} & \textbf{circuit\_logic (OOD)} \\
\midrule
\multirow{9}{*}{\textbf{ACC ($\uparrow$)}} & MLE & 0.848$_{\pm 0.009}$ & 0.660$_{\pm 0.069}$ & 0.841$_{\pm 0.008}$ & 0.613$_{\pm 0.000}$ \\
 & MAP & 0.847$_{\pm 0.004}$ & 0.659$_{\pm 0.061}$ & 0.840$_{\pm 0.005}$ & 0.613$_{\pm 0.000}$ \\
 & TempScale & 0.848$_{\pm 0.009}$ & 0.660$_{\pm 0.069}$ & 0.841$_{\pm 0.008}$ & 0.613$_{\pm 0.000}$ \\
 & MCDropout & 0.850$_{\pm 0.009}$ & 0.641$_{\pm 0.023}$ & 0.841$_{\pm 0.003}$ & 0.613$_{\pm 0.000}$ \\
 & Ensemble & 0.835$_{\pm 0.008}$ & 0.781$_{\pm 0.087}$ & 0.842$_{\pm 0.002}$ & 0.613$_{\pm 0.001}$ \\
 & Laplace & 0.854$_{\pm 0.001}$ & 0.658$_{\pm 0.067}$ & 0.843$_{\pm 0.012}$ & 0.613$_{\pm 0.000}$ \\
\bottomrule
\end{tabular}
\label{tab:mmlu_clogic_4B}
\end{table*}

\begin{table*}[h!]
\centering
\caption{OOD Performance Comparison for Qwen3-8B}
\begin{tabular}{@{}cccccc@{}}
\toprule
\textbf{Metric} & \textbf{Method} & \multicolumn{2}{c}{\textbf{Train on circuit\_logic}} & \multicolumn{2}{c}{\textbf{Train on expression\_logic}} \\
\cmidrule(lr){3-4} \cmidrule(lr){5-6}
 &  & \textbf{In-dist} & \textbf{expression\_logic (OOD)} & \textbf{In-dist} & \textbf{circuit\_logic (OOD)} \\
\midrule
\multirow{9}{*}{\textbf{ACC ($\uparrow$)}} & MLE & 0.842$_{\pm 0.003}$ & 0.823$_{\pm 0.010}$ & 0.840$_{\pm 0.008}$ & 0.645$_{\pm 0.029}$ \\
 & MAP & 0.840$_{\pm 0.006}$ & 0.826$_{\pm 0.007}$ & 0.837$_{\pm 0.008}$ & 0.668$_{\pm 0.058}$ \\
 & TempScale & 0.842$_{\pm 0.003}$ & 0.823$_{\pm 0.010}$ & 0.840$_{\pm 0.008}$ & 0.645$_{\pm 0.029}$ \\
 & MCDropout & 0.842$_{\pm 0.004}$ & 0.806$_{\pm 0.022}$ & 0.838$_{\pm 0.003}$ & 0.669$_{\pm 0.063}$ \\
 & Ensemble & 0.824$_{\pm 0.002}$ & 0.825$_{\pm 0.011}$ & 0.843$_{\pm 0.002}$ & 0.652$_{\pm 0.006}$ \\
 & Laplace & 0.840$_{\pm 0.000}$ & 0.824$_{\pm 0.010}$ & 0.840$_{\pm 0.002}$ & 0.641$_{\pm 0.001}$ \\
\bottomrule
\end{tabular}
\label{tab:mmlu_clogic_8B}
\end{table*}

\begin{table*}[h!] 
\centering
\caption{Gemma versus Qwen on Winogrande Small In-Distribution}
\begin{tabular}{@{}cccc@{}}
\toprule
\textbf{Metric} & \textbf{Method} & \textbf{gemma-3-4b-it} & \textbf{Qwen3-4B} \\
\midrule
\multirow{4}{*}{\textbf{ACC ($\uparrow$)}} & MLE & 0.681$_{\pm 0.002}$ & 0.711$_{\pm 0.004}$ \\
 & MCDropout & 0.684$_{\pm 0.003}$ & 0.709$_{\pm 0.004}$ \\
 & BLoB & 0.682$_{\pm 0.005}$ & 0.727$_{\pm 0.005}$ \\
 & ScalaBL & 0.663$_{\pm 0.003}$ & 0.716$_{\pm 0.007}$ \\
\midrule
\multirow{4}{*}{\textbf{ECE ($\downarrow$)}} & MLE & 0.311$_{\pm 0.003}$ & 0.268$_{\pm 0.004}$ \\
 & MCDropout & 0.298$_{\pm 0.003}$ & 0.270$_{\pm 0.003}$ \\
 & BLoB & 0.110$_{\pm 0.011}$ & 0.099$_{\pm 0.011}$ \\
 & ScalaBL & 0.085$_{\pm 0.003}$ & 0.110$_{\pm 0.005}$ \\
\midrule
\multirow{4}{*}{\textbf{NLL ($\downarrow$)}} & MLE & 4.182$_{\pm 1.204}$ & 2.789$_{\pm 0.202}$ \\
 & MCDropout & 3.562$_{\pm 0.871}$ & 2.773$_{\pm 0.330}$ \\
 & BLoB & 0.636$_{\pm 0.015}$ & 0.591$_{\pm 0.021}$ \\
 & ScalaBL & 0.633$_{\pm 0.003}$ & 0.621$_{\pm 0.010}$ \\
\bottomrule
\end{tabular}
\label{tab:gemma_vs_qwen_id_ws}
\end{table*}
\begin{table*}[h!] 
\centering
\caption{Gemma versus Qwen on SLAKE In-Distribution} 
\begin{tabular}{@{}cccc@{}}
\toprule
\textbf{Metric} & \textbf{Method} & \textbf{gemma-3-4b-it} & \textbf{Qwen3-VL-4B-Instruct} \\
\midrule
\multirow{4}{*}{\textbf{ACC ($\uparrow$)}} & MLE & 0.875$_{\pm 0.006}$ & 0.894$_{\pm 0.005}$ \\
 & MCDropout & 0.865$_{\pm 0.017}$ & 0.899$_{\pm 0.004}$ \\
 & BLoB & 0.875$_{\pm 0.011}$ & 0.900$_{\pm 0.006}$ \\
 & ScalaBL & 0.840$_{\pm 0.005}$ & 0.891$_{\pm 0.007}$ \\
\midrule
\multirow{4}{*}{\textbf{ECE ($\downarrow$)}} & MLE & 0.116$_{\pm 0.006}$ & 0.100$_{\pm 0.004}$ \\
 & MCDropout & 0.123$_{\pm 0.012}$ & 0.095$_{\pm 0.004}$ \\
 & BLoB & 0.035$_{\pm 0.006}$ & 0.032$_{\pm 0.008}$ \\
 & ScalaBL & 0.045$_{\pm 0.010}$ & 0.035$_{\pm 0.008}$ \\
\midrule
\multirow{4}{*}{\textbf{NLL ($\downarrow$)}} & MLE & 0.742$_{\pm 0.054}$ & 0.866$_{\pm 0.047}$ \\
 & MCDropout & 0.675$_{\pm 0.034}$ & 0.856$_{\pm 0.147}$ \\
 & BLoB & 0.292$_{\pm 0.003}$ & 0.240$_{\pm 0.007}$ \\
 & ScalaBL & 0.350$_{\pm 0.004}$ & 0.258$_{\pm 0.010}$ \\
\bottomrule
\end{tabular}
\label{tab:gemma_vs_qwen_id_slake}
\end{table*}

\end{document}